\documentclass[journal,twoside,web]{ieeecolor}
\usepackage{tmi}
\usepackage{amsmath,amssymb}
\usepackage{algorithmic}
\usepackage{graphicx}
\usepackage{color}
\usepackage{booktabs}
\usepackage{makecell}
\usepackage{multirow}
\usepackage{mathrsfs}
\usepackage{subcaption}
\usepackage{rotating}
\usepackage{balance}
\usepackage{cite}
\usepackage{ulem}

\newcommand{\eg}{\textit{e.g.}}

\newcommand{\etal}{\textit{et al.}}

\DeclareGraphicsExtensions{.pdf,.png,.jpg,.fig,.tif}
\usepackage[linesnumbered,ruled]{algorithm2e}

\definecolor{color1}{rgb}{0.2353,0.2353,1}
\definecolor{color2}{rgb}{0.8000,0,0.4000}
\definecolor{color3}{rgb}{0.6980,0.4000,1}
\definecolor{color4}{rgb}{1,0.6,1}

\newcommand{\alexnet}{{AlexNet }}
\newcommand{\vgg}{VGG}
\newcommand{\resnet}{{ResNet}}
\newcommand{\densenet}{{DenseNet }}
\newcommand{\googlenet}{{GoogLeNet }}

\newcommand{\mobilenet}{{MobileNet}}

\newcommand{\inception}{{Inception}}
\newcommand{\efficientnet}{{EfficientNet }}

\newcolumntype{L}[1]{>{\raggedright\arraybackslash}p{#1}}
\newcolumntype{C}[1]{>{\centering\arraybackslash}p{#1}}
\newcolumntype{R}[1]{>{\raggedleft\arraybackslash}p{#1}}

\def\BibTeX{{\rm B\kern-.05em{\sc i\kern-.025em b}\kern-.08em
    T\kern-.1667em\lower.7ex\hbox{E}\kern-.125emX}}
\begin{document}
\title{One-Vote Veto: Semi-Supervised Learning for Low-Shot Glaucoma Diagnosis}
\author{Rui Fan, \IEEEmembership{Senior Member, IEEE}, Christopher Bowd, Nicole Brye, Mark Christopher,\\ Robert N. Weinreb, David J. Kriegman, \IEEEmembership{Fellow, IEEE}, Linda M. Zangwill
\thanks{This research was supported in part by the Fundamental Research Funds for the Central Universities, and was also supported in part by awards from the National Eye Institute (grants EY027510, R214278211, P30EY022589, K99EY030942, EY026574, EY11008, and EY19869), the National Center on Minority Health and Health Disparities, National Institutes of Health (grants EY09341 and EY09307), Horncrest Foundation, awards to the Department of Ophthalmology and Visual Sciences at Washington University, the NIH Vision Core Grant P30 EY 02687, Merck Research Laboratories, Pfizer, Inc., White House Station, New Jersey, and unrestricted grants from Research to Prevent Blindness, Inc., New York, NY. \textit{(Corresponding author: Linda M. Zangwill)}}
\thanks{R. Fan is with the College of Electronics \& Information Engineering, Shanghai Research Institute for Intelligent Autonomous Systems, State Key Laboratory of Intelligent Autonomous Systems, and Frontiers Science Center for Intelligent Autonomous Systems, Tongji University, Shanghai 201804, China, and was with Hamilton Glaucoma Center, Viterbi Family Department of Ophthalmology, Shiley Eye Institute, and the Department of Computer Science \& Engineering, the University of California San Diego, La Jolla, CA 92093, USA 
	(e-mail: rui.fan@ieee.org).}
\thanks{C. Bowd, N. Brye, M. Christopher, R. N. Weinreb, and L. M. Zangwill are with Hamilton Glaucoma Center, Viterbi Family Department of Ophthalmology and Shiley Eye Institute, the University of California San Diego, La Jolla, CA 92093, USA (e-mail: \{cbowd, nbrye, mac157, rweinreb,  lzangwill\}@health.ucsd.edu).}
\thanks{D. J. Kriegman is with the Department of Computer Science \& Engineering, the University of California San Diego, La Jolla, CA 92093, USA (e-mail: kriegman@ucsd.edu).}
}
\maketitle
\begin{abstract}
Convolutional neural networks (CNNs) are a promising technique for automated glaucoma diagnosis from images of the fundus, and these images are routinely acquired as part of an ophthalmic exam. Nevertheless, CNNs typically require a large amount of well-labeled data for training, which may not be available in many biomedical image classification applications, especially when diseases are rare and where labeling by experts is costly. This article makes two contributions to address this issue: (1) It extends the conventional Siamese network and introduces a training method for low-shot learning when labeled data are limited and imbalanced, and (2) it introduces a novel semi-supervised learning strategy that uses additional unlabeled training data to achieve greater accuracy. Our proposed multi-task Siamese network (MTSN) can employ any backbone CNN, and we demonstrate with four backbone CNNs that its accuracy with limited training data approaches the accuracy of backbone CNNs trained with a dataset that is 50 times larger. We also introduce One-Vote Veto (OVV) self-training, a semi-supervised learning strategy that is designed specifically for MTSNs. By taking both self-predictions and contrastive predictions of the unlabeled training data into account, OVV self-training provides additional pseudo labels for fine-tuning a pre-trained MTSN. Using a large (imbalanced) dataset with 66,715 fundus photographs acquired over 15 years, extensive experimental results demonstrate the effectiveness of low-shot learning with MTSN and semi-supervised learning with OVV self-training. Three additional, smaller clinical datasets of fundus images acquired under different conditions (cameras, instruments, locations, populations) are used to demonstrate the generalizability of the proposed methods. 
\end{abstract}

\section{Introduction}
\label{sec.introduction}

\IEEEPARstart{G}laucoma is a prevalent and debilitating disease that can lead to progressive and irreversible vision loss through optic nerve damage \cite{weinreb2004primary}. The global incidence of glaucoma was estimated at 64.3 million in 2013, and due to aging populations, this number is expected to rise to 111.8 million by 2040 \cite{tham2014global}. Improvement in the management of glaucoma would have a major human and socio-economic impact \cite{fan2022detecting}. Early identification and intervention would significantly reduce the economic burden of late-stage disease  \cite{traverso2005direct}. In addition, visual impairment in glaucoma patients has been associated with decreased physical activity and mental health \cite{huang2020adverse,parrish1997visual} and increased risk of motor vehicle accidents \cite{kwon2016association, mcgwin2015binocular}.

\begin{figure}[!t]
	\begin{center}
		\centering
		\includegraphics[width=0.485\textwidth]{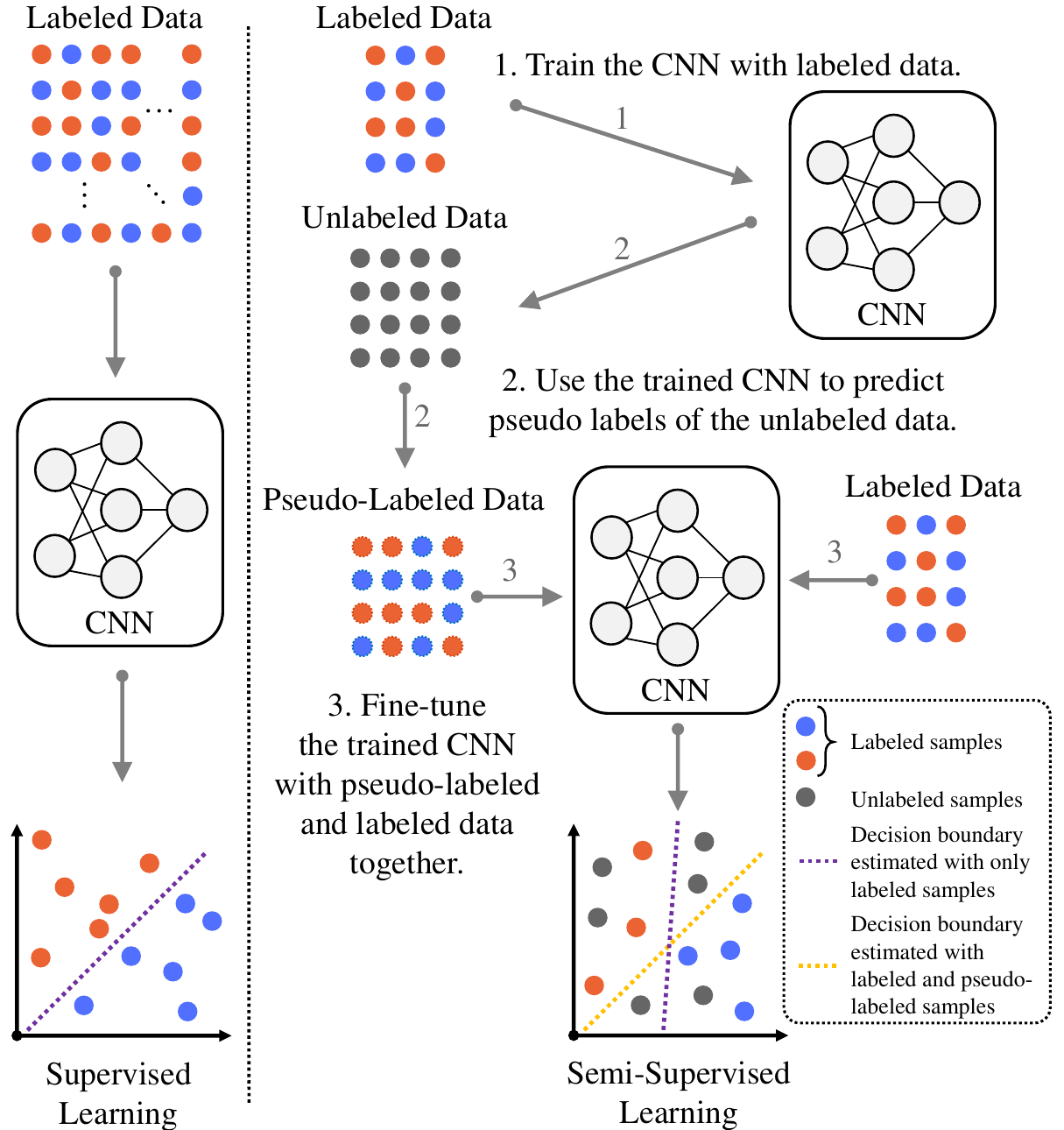}
		\centering
		\caption{Supervised learning v.s. semi-supervised learning.}
		\label{fig.sl_vs_ssl}
	\end{center}
\end{figure}

With the recent advances in machine learning, convolutional neural networks (CNNs), trained via supervised learning, have shown promise in diagnosing glaucoma from fundus images (photographs of the back of eyes) \cite{lag}. However, this requires large amounts of empirical data for supervised training \cite{fan2023transformer}. {In this study, we use 66,715 fundus photographs from the \uline{\textbf{O}cular \textbf{H}ypertension \textbf{T}reatment \textbf{S}tudy (\textbf{OHTS})} \cite{ohts1, ohts2, ohts3}}, which is a 22-site multi-center, longitudinal (phase 1 and 2, 1994-2008) randomized clinical trial of 1,636 subjects (3,272 eyes). The primary goal of the OHTS was to determine if topical ocular hypotensive medications could delay or prevent the onset of glaucoma in eyes with high intraocular pressure \cite{ohts1}. Conversion to glaucoma was decided by a masked endpoint committee of three glaucoma specialists using fundus photographs and visual fields. Owing to its well-characterized ground-truth labels, the OHTS dataset provides us a basis to explore an effective way of training CNNs to diagnose glaucoma with low-shot learning when only a small quantity of labeled data is available, and/or semi-supervised learning when raw data is abundant, but labeling resources are scarce, costly, require strong expertise, or are just unavailable. However, as shown in Fig. \ref{fig.sl_vs_ssl}, conventional semi-supervised learning approaches typically require a reliable pre-trained CNN (using a small sample) as prior knowledge, which is often challenging due to the over-fitting problem. Moreover, there is also a strong motivation to design a {feasible semi-supervised learning strategy capable} of determining confident predictions and generating pseudo labels for unlabeled data. We focus specifically on fundus images and glaucoma diagnosis in this article because we have sufficient data to accurately characterize the effectiveness of our methods. The same techniques could also be applied to tasks where there is limited data, such as rare diseases or where limited labels are available (\eg, asthma and diabetes prediction from fundus images). Therefore, this article aims to answer the following questions:
\begin{enumerate}
	\item Can a CNN be developed to accurately diagnose glaucoma, compared to the expert graders of the OHTS? Will the model be generalizable to other datasets?
	\item Is it necessary to train CNNs with thousands of labeled fundus images to diagnose glaucoma, or can diagnosis be achieved using only one image per patient (approximately 1.1K fundus images in the OHTS training set)?
	\item Can the performance of a CNN trained using a small sample be improved further by fine-tuning it with additional unlabeled training data?
\end{enumerate}

To answer these questions, we first evaluate the performance of state-of-the-art (SoTA) glaucoma diagnosis algorithms, including six supervised learning algorithms \cite{li2018efficacy, gomez2019automatic, judy2020automated, serener2019transfer, thakur2020predicting, fan2022detecting}, one low-shot learning algorithm \cite{kim2017few}, and two semi-supervised learning algorithms \cite{al2019semi, diaz2019retinal}, on the OHTS dataset. Their generalizabilities are further validated on three additional clinical datasets of fundus images: (a) ACRIMA (Spain) \cite{acrima}, (b) Large-Scale Attention-Based Glaucoma (LAG, China) \cite{lag}, and (c) the UCSD-based Diagnostic Innovations in Glaucoma Study and African Descent and Glaucoma Evaluation Study (DIGS/ADAGES, US) \cite{digs}. 

\begin{figure}[!t]
	\begin{center}
		\centering
		\includegraphics[width=0.49\textwidth]{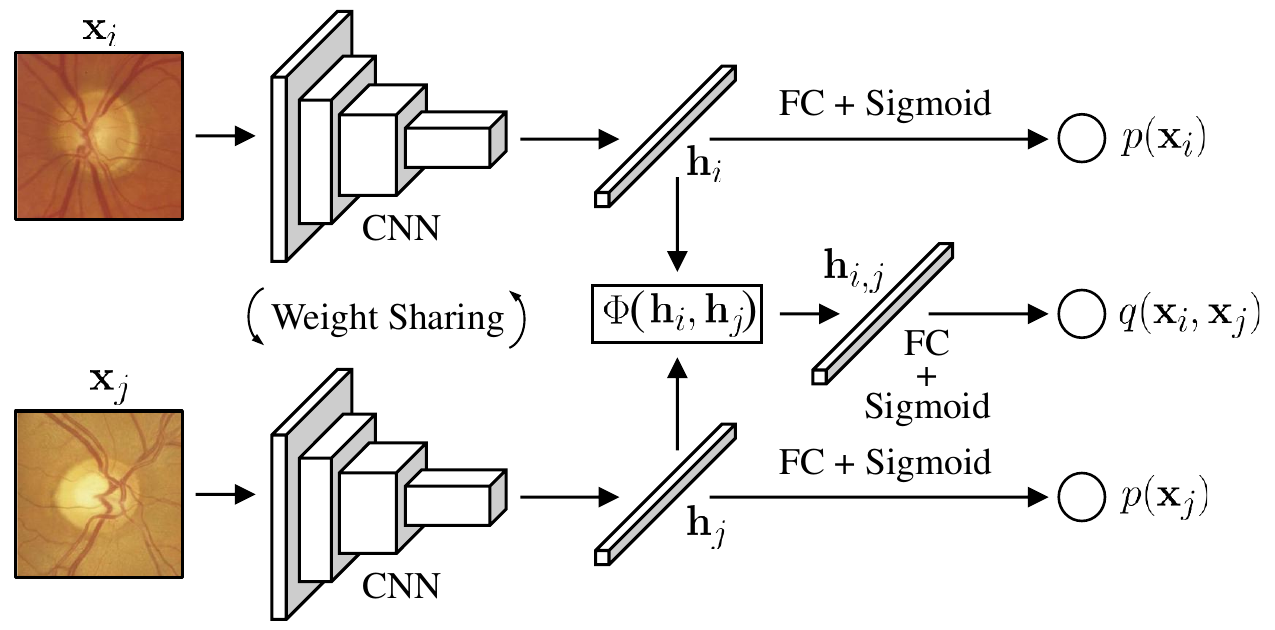}
		\centering
		\caption{
		An illustration of our MTSN for joint learning of fundus image similarity measurement (primary task) and glaucoma diagnosis (secondary task) in a low-shot manner.}
		\label{fig.twin_network}
	\end{center}
\end{figure}

Furthermore, we propose a novel extension to the conventional Siamese network, referred to as the \uline{\textbf{M}ulti-\textbf{T}ask \textbf{S}iamese \textbf{N}etwork (\textbf{MTSN})}, as depicted in Fig. \ref{fig.twin_network}. By minimizing a novel \uline{\textbf{C}ombined \textbf{W}eighted \textbf{C}ross-\textbf{E}ntropy (\textbf{CWCE}) \textbf{Loss}}, the MTSN can simultaneously perform two tasks: measuring the similarity of a given pair of images (primary task) and classifying them as healthy or glaucoma (secondary task). With a small training set of approximately 1.1K fundus images, we explore the feasibility of training an MTSN for glaucoma diagnosis. Although the MTSN may not provide complementary information, it effectively performs a type of ``data augmentation'' by generating $C(N,2)$ pairs of fundus images for training instead of using $N$ independent fundus images. The visual features learned from these two tasks prove to be more informative for glaucoma diagnosis when the training set is small. Our experimental results demonstrate that the MTSN greatly reduces over-fitting and achieves an accuracy on a small training set comparable to a large training set, which contains approximately 53K fundus images.

Moreover, we propose a novel semi-supervised learning strategy, referred to as \uline{\textbf{O}ne-\textbf{V}ote \textbf{V}eto (\textbf{OVV}) \textbf{Self-Training}}, which generates reliable pseudo labels for the unlabeled training data and incorporates them into the labeled training data to fine-tune the MTSN for improved performance and generalizability. Our extensive experiments show that the MTSN fine-tuned with OVV self-training achieves similar performance to the corresponding backbone CNN trained via supervised learning on the OHTS dataset, and achieves higher area under the receiver operating characteristic curve (AUROC) scores on the additional fundus image datasets. The fine-tuned MTSN also outperforms SoTA semi-supervised glaucoma diagnosis approaches \cite{al2019semi, diaz2019retinal}, and in some cases, even outperforms SoTA supervised approaches. Additionally, we compare our proposed OVV self-training approach with four SoTA general-purpose semi-supervised learning methods, including FreeMatch \cite{wang2022freematch}, SoftMatch \cite{chen2023softmatch}, FixMatch \cite{sohn2020fixmatch}, and FlexMatch \cite{zhang2021flexmatch}, all of which utilize vision Transformer \cite{dosovitskiyimage} as their backbone network. The results demonstrate that our proposed OVV self-training approach outperforms these methods on the OHTS dataset and demonstrates better generalizability on three additional fundus image test sets.

We also conduct two additional few-shot biomedical image classification experiments (chest X-ray image classification \cite{chowdhury2020can, rahman2020exploring} and lung histopathological image classification \cite{borkowski2019lung}) to further validate the effectiveness of the MTSN on other types of image data. The promising results indicate that our proposed algorithms have the potential to solve a variety of biomedical image classification problems. 

\section{Related Works}
\label{sec.related_works}
Most SoTA glaucoma diagnosis algorithms are developed based on supervised fundus image classification. For example, Judy \etal \cite{judy2020automated} trained an \alexnet \cite{alexnet} to diagnose glaucoma. As {\vgg} architectures \cite{vgg} can learn more complicated image features than AlexNet, G{\'o}mez-Valverde \etal \cite{gomez2019automatic} employed a \vgg-19 \cite{vgg} model to diagnose glaucoma. Nevertheless, {\vgg} architectures \cite{vgg} consist of hundreds of millions of parameters, making them very memory-consuming. In contrast, \googlenet \cite{googlenet} and \inception-v3 \cite{inception} have lower computational complexities. Hence, Ahn \etal \cite{ahn2018deep} and Li \etal \cite{li2018efficacy} utilized transfer learning to re-train an \inception-v3 \cite{inception} model (pre-trained on the ImageNet \cite{imagenet_cvpr09} database) for glaucoma diagnosis, while Serener and Serte \cite{serener2019transfer} re-trained a pre-trained \googlenet \cite{googlenet} model to diagnose glaucoma. However, with the increase of network depth, accuracy gets saturated and then degrades rapidly due to vanishing gradients \cite{resnet}. To tackle this problem, the residual neural network ({\resnet}) \cite{resnet} was developed. Due to its robustness, \resnet-50 \cite{resnet} has been extensively used for biomedical image analysis, and it is a popular choice \cite{liu2018deep, ran2019detection, medeiros2020detection, christopher2018performance, christopher2020effects, fan2022detecting} for fundus image classification. Additionally, developing low-cost and real-time embedded glaucoma diagnosis systems \cite{jain2016open, matthew2014smart, thakur2020predicting}, \eg, based on \mobilenet-v2 \cite{mobilenetv2}, for mobile devices is also an emerging area.

Machine/deep learning has achieved compelling performance in data-intensive applications, but it is often challenging for these algorithms to yield comparable performance when only a limited amount of labeled training data is available \cite{fslsurvey}. Low-shot and semi-supervised learning can address these issues.  Unfortunately, they are rarely discussed in the field of glaucoma diagnosis. To the best of our knowledge, \cite{kim2017few} is the only published low/few-shot glaucoma diagnosis algorithm. This algorithm employs a conventional Siamese network to compare two groups of (negative and positive) fundus images. The Siamese network utilizes two identical CNNs to learn visual embeddings. A bi-directional long short-term memory \cite{zhou2016attention} component is then trained over the CNN outputs for glaucoma diagnosis. However, the training process is complicated since different types of losses are minimized, and the achieved glaucoma diagnosis results are unsatisfactory since each sub-network is only fed with one type of fundus images (either negative or positive). The lack of same-class comparisons leads to a performance bottleneck when compared to the MTSN proposed in this article. 

A thorough search of the relevant literature yielded only two published studies on semi-supervised learning specifically for glaucoma diagnosis \cite{diaz2019retinal, al2019semi}. Diaz-Pinto \etal \cite{diaz2019retinal} utilized a deep convolutional generative adversarial network (DCGAN) \cite{radford2015unsupervised} for semi-supervised learning of glaucoma diagnosis, where the discriminator is trained to classify healthy and glaucomatous optic neuropathy (GON) fundus images, while also distinguishing between real and fake fundus images. The classifier for the former task is then employed for glaucoma diagnosis. On the other hand, Al Ghamdi \etal \cite{al2019semi} developed a glaucoma diagnosis approach based on self-training \cite{selftraining}, which is a typical semi-supervised learning approach that uses a pre-trained model (typically yielded via supervised learning) to produce pseudo labels of the unlabeled data. However, producing reliable pseudo labels is a significant challenge in self-training, and the pseudo labels generated by a single pre-trained CNN are usually not trustworthy enough for CNN fine-tuning \cite{van2020survey}. Additionally, training a reliable pre-trained classifier with only a small amount of labeled data is notably demanding. In this article, we combine semi-supervised learning with low-shot learning to address these issues using glaucoma diagnosis as an example case. Specifically, our proposed OVV self-training strategy, as discussed in Sect. \ref{sec.ovv}, is inspired by the mechanism of \textit{learning with external memory} (LwEM), used in low-shot learning \cite{miller2016key}, where the labels of unlabeled training data are predicted by a classifier trained via low-shot learning on a small collection of fundus images with ground-truth labels.

\section{Methodology}
\label{sec.methodology}

\subsection{Multi-Task Siamese Network}
\label{sec.tnn}

As illustrated in Fig. \ref{fig.sl_vs_ssl}, conventional semi-supervised learning methods initialize a network by pre-training it with a small number of fundus images for glaucoma diagnosis. However, we observed that such approaches are highly sensitive to noise. As a result, we design a novel MTSN specifically for our semi-supervised learning approach, which requires not only predicting the label of a given fundus image but also determining the similarity between a pair of given fundus images to generate pseudo labels through a voting process.

Conventional Siamese networks have become a common choice for metric learning and few/low-shot image recognition tasks \cite{siamese}. These networks comprise two identical sub-networks, as depicted in Fig. \ref{fig.twin_network}. Each pair of fundus images $\mathbf{x}_i$ and $\mathbf{x}_j$ are separately fed into these sub-networks, producing two 1D embeddings (features) $\mathbf{h}_i$ and $\mathbf{h}_j$, respectively. Another 1D embedding $\mathbf{h}_{i,j}$ is generated by $\Phi(\cdot)$. $\mathbf{h}_{i,j}$ is then passed through a fully connected (FC) layer to produce a scalar $q(\mathbf{x}_i,\mathbf{x}_j)\in[0,1]$ indicating the similarity between $\mathbf{x}_i$ and $\mathbf{x}_j$. If $\mathbf{x}_i$ and $\mathbf{x}_j$ are dissimilar, $q(\mathbf{x}_i,\mathbf{x}_j)$ approaches 1, and vice versa. The ground-truth labels of $\mathbf{x}_i$ and $\mathbf{x}_j$ are represented by $y_i\in\{0,1\}$ and $y_j\in\{0,1\}$, respectively, where 0 denotes a healthy image, and 1 denotes a GON image.

However, a conventional Siamese network can only determine whether $\mathbf{x}_i$ and $\mathbf{x}_j$ belong to the same category, rather than predicting their independent categories. A straightforward solution is to connect $\mathbf{h}_i$ and $\mathbf{h}_j$ to separate FC layers, producing two scalars $p(\mathbf{x}_i)$ and $p(\mathbf{x}_j)$ indicating the probabilities that $\mathbf{x}_i$ and $\mathbf{x}_j$ are GON images, respectively. Refer to Fig. \ref{fig.twin_network} and note that the two FC layers connected to $\mathbf{h}_i$ and $\mathbf{h}_j$ use the same weights. In this article, we refer to the network architecture in Fig. \ref{fig.twin_network} as an MTSN, which can simultaneously measure the similarity of a given pair of fundus images and classify them as either healthy or GON. These two tasks are dependent yet not directly deducible from one another. A well-trained glaucoma diagnosis network can be employed to compare differences between given pairs of fundus images, but a well-trained fundus image similarity measurement network cannot directly output the category of a given fundus image.

In addition, the visual features learned from the primary and secondary tasks are distinct from one another. For the primary task, the network learns the visual features to classify same-class and different-class fundus image pairs. On the other hand, for the secondary task, the network learns the visual features to classify GON and healthy fundus images. Although this network architecture may not provide complementary information, it effectively performs a type of ``data augmentation'' by producing $C(N,2)$ pairs of fundus images for training, rather than using $N$ independent fundus images. The visual features learned from these two tasks prove to be more informative for glaucoma diagnosis when the training set is small. Furthermore, multi-task learning is effective because requiring an algorithm to perform well on a related task induces regularization, which can be superior to uniform complexity penalization for preventing over-fitting. This idea has been explored in many Siamese neural network works, such as \cite{jang2021siamese, wang2018learning, wang2015face}.

\begin{figure*}[!t]
	\begin{center}
		\centering
		\includegraphics[width=0.99999\textwidth]{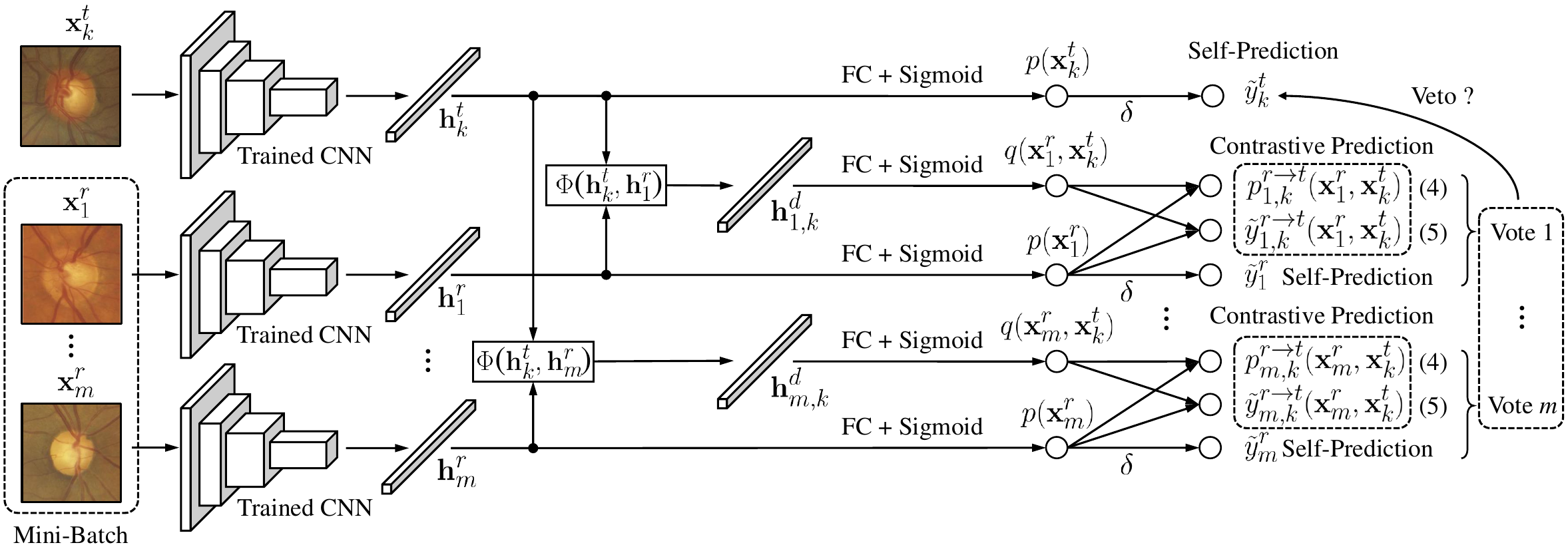}
		\centering
		\caption{
			An illustration of our One-Vote Veto Self-Training strategy. $\mathbf{h}^d_{1,k}$ and $\mathbf{h}^d_{m,k}$ are two 1D embeddings, followed by an FC layer to produce scalars indicating the similarities between the given pairs of reference and target fundus images. The reference fundus images having ground-truth labels are used to train the MTSN by minimizing (\ref{eq.L}). The contrastive predictions are obtained using (\ref{eq.con_prob}) and (\ref{eq.con_label}). The pseudo labels of the target fundus photographs are generated using One-Vote Veto Self-Training strategy, as detailed in Algorithm \ref{alg.ovv}. 
		}
		\label{fig.ovv}
	\end{center}
\end{figure*}

In this article, we use $n_0$ and $n_1$ to respectively denote the numbers of healthy and GON fundus images used to train the MTSN, with $n=n_0+n_1$. $n_0$ is usually much greater than $n_1$, because there are fewer patients with glaucomatous disease than healthy patients, resulting in a severely imbalanced dataset. Therefore, the MTSN is trained by minimizing a CWCE loss as follows: 
\begin{equation}
	\mathcal{L}=\mathcal{L}_\text{sim}+\lambda\mathcal{L}{_{\text{cla}}},
	\label{eq.L}
\end{equation}
where
\begin{align}
	\begin{split}
		\mathcal{L}_\text{sim}&= - \frac{n_0(n_0-1)+ n_1(n_1-1)}{n(n-1)}|y_i-y_j|\log(q(\mathbf{x}_i,\mathbf{x}_j))\\ &-\frac{2n_0n_1}{n(n-1)}(1-|y_i-y_j|)\log(1-q(\mathbf{x}_i,\mathbf{x}_j)) ,
		\label{eq.L_sim}
	\end{split}
\end{align}
\begin{align}
	\begin{split}
		&\mathcal{L}{_{\text{cla}}}= - \frac{1}{n}\Big(
		n_0\big(
		y_{i}\log(p(\mathbf{x}_i))+
		y_{j}\log(p(\mathbf{x}_j)) \big)
		\\&+n_1\big((1-y_i)\log(1-p(\mathbf{x}_i))+(1-y_j)\log(1-p(\mathbf{x}_j)) \big)
		\Big), 
		\label{eq.L_cla}
	\end{split}
\end{align}
The hyper-parameter $\lambda$ balances the primary task loss $\mathcal{L}_{\text{sim}}$ and the secondary task loss $\mathcal{L}_{\text{cla}}$. The choice of $\lambda$ and $\Phi(\cdot)$ is discussed in Sect. \ref{sec.exp_fsl}.
The motivations for using such a CWCE loss function instead of the commonly used triplet loss \cite{triplet} or contrastive loss \cite{khosla2020supervised} to train the MTSN are:

\begin{enumerate}
	\item Most datasets for rare disease diagnosis are imbalanced. As detailed in Sect. \ref{sec.exp_datasets}, the OHTS training set is severely imbalanced, with 50,208 healthy images and only 2,416 GON images for supervised learning, and 995 healthy images and 152 GON images for low-shot learning. Learning from such an imbalanced dataset without weights on different classes can result in many incorrect predictions, with most GON images likely to be predicted as healthy images. To address this issue, a higher weight should be used for the minority class to prevent the CNN from predicting all fundus images as the majority class.
	\item In multi-task learning, weighing different types of losses, such as regression and classification, is typically challenging \cite{kendall2018multi}. {Assigning an incorrect weight may cause one task to perform poorly, even when other tasks converge to satisfactory results.} Therefore, formulating  $\mathcal{L}_\text{sim}$ as a weighted cross-entropy loss function is a simple but effective solution. However, due to the dataset imbalance problem, the cross-entropy losses have to be weighted.
	\item As shown in Fig. \ref{fig.ovv}, OVV self-training requires both labels and probabilities (of being GON images), predicted by a pre-trained model, to produce pseudo labels for unlabeled data. Such network architecture and training loss can efficiently and effectively provide both ``self-predicted'' and ``contrastively-predicted'' labels and probabilities, as described in Sect. \ref{sec.ovv}. 
\end{enumerate}
It should be noted here that using the fundus images from the same patient as an image pair for MTSN training is not necessary. 

\subsection{One-Vote Veto Self-Training}
\label{sec.ovv}

As discussed in Sect. \ref{sec.introduction}, self-training aims to improve the performance of a pre-trained model by incorporating reliable predictions of the unlabeled data to obtain useful additional information that can be used for model fine-tuning. A feasible strategy to determine such reliable predictions is, therefore, key to the success of self-training \cite{mcclosky2006effective}.

In conventional semi-supervised learning algorithms, a pre-trained image classification model (obtained through supervised learning) can be fine-tuned by assessing the reliability of unlabeled images. To determine the reliability of an unlabeled image, its probability distribution for the most likely class is compared to a pre-determined threshold. If the probability surpasses this threshold, the prediction is considered a pseudo label. Subsequently, the image and its corresponding pseudo label are utilized to fine-tune the pre-trained model.

However, relying {solely} on probability distributions to generate pseudo labels is often insufficient \cite{selftraining}. Drawing inspiration from LwEM \cite{sukhbaatar2015end}, we introduce \textit{One-Vote Veto self-training} in this paper, as illustrated in Fig. \ref{fig.ovv}\footnote{The superscripts  $r$ and $t$ denote ``reference'' and ``target'', respectively.}. Similar to LwEM \cite{sukhbaatar2015end}, we use a collection of $m$ reference (labeled) fundus images $\{ \mathbf{x}^r_1,\dots,\mathbf{x}^r_m  \}\in \mathscr{X}^r$ to provide ``\textit{contrastive predictions}'' to the target (unlabeled) fundus images $\{ \mathbf{x}^t_1,\dots,\mathbf{x}^t_m  \}\in \mathscr{X}^t$. The contrastive predictions subsequently vote to veto the unreliable ``\textit{self-predictions}'' $\{ \tilde{y}^t_1, \dots, \tilde{y}^t_m   \}$ produced by the MTSN. Our OVV self-training is detailed in Algorithm \ref{alg.ovv}, where the target model updates its parameters during self-training but the reference model does not. 

\begin{algorithm}[t!]
	{
		\fontsize{9.0}{9.7}\selectfont
		\KwData{Reference fundus images $\mathscr{X}^r$ and their labels $\mathscr{Y}^r$, and target fundus images $\mathscr{X}^t$ }
		\While{Training}{
			Given a mini-batch consisting of $\{\mathbf{x}^r_1,\dots,\mathbf{x}^r_m\}\in \mathscr{X}^r, \  \{y^r_1,\dots,y^r_m\} \in \mathscr{Y}^r$ and
			$\{\mathbf{x}^t_1,\dots,\mathbf{x}^t_m\} \in \mathscr{X}^t$\;
			
			{Initialize an empty set $\mathscr{P}$ to store reliable target fundus images and their pseudo labels \;}
			
			\For{{Each target fundus image $\mathbf{x}^t_k$}}{		
				\If{{The self-prediction of $\mathbf{x}^t_k$ is reliable}}{
					\For{{Any qualified reference image $\mathbf{x}^r_l$}}{	
						{Compute the self-prediction and contrastive prediction}\;
					}
					
					\If{{The criteria to generate pseudo labels are satisfied}
					}{
						{Update the set $\mathscr{P}$}\;
					}	
				}	
			}
			Fine-tune the target model using {unique($\mathscr{P}$)}\;
		}	
		\If{The target model outperforms the reference model}{Update the reference model parameters\;}}			
	\caption{\textbf{One-Vote Veto Self-Training Strategy.}}
	\label{alg.ovv}	
\end{algorithm}

When fine-tuning an MTSN pre-trained through low-shot learning, each mini-batch contains a discrete set of $m$ reference fundus images $\{ \mathbf{x}^r_1,\dots,\mathbf{x}^r_m  \}\in \mathscr{X}^r$, their ground-truth labels $\{ {y}^r_1,\dots,{y}^r_m  \}\in \mathscr{Y}^r$, and {an equal} number of $m$ target fundus images $\{ \mathbf{x}^t_1,\dots,\mathbf{x}^t_m  \}\in \mathscr{X}^t$ without labels. $\mathbf{h}^r_k$ and $\mathbf{h}^t_k$ represent the 1D embeddings learned from $\mathbf{x}^r_k$ and $\mathbf{x}^t_k$ ($k\in[1,m]\cap\mathbb{Z}$), respectively. Given a pair of reference and target fundus images, $\mathbf{x}^r_l$ and $\mathbf{x}^t_k$, the pre-trained MTSN can ``self-predict'': 
\begin{itemize}
\item the scalars ${p}(\mathbf{x}^r_l)$ and ${p}(\mathbf{x}^t_k)$ which indicate the probabilities that $\mathbf{x}^r_l$ and $\mathbf{x}^t_k$ are GON images, respectively;
\item their labels $\tilde{y}^r_l=\delta({{p}(\mathbf{x}^r_l)})$ and $\tilde{y}^t_k=\delta({{p}(\mathbf{x}^t_k)})$ using its fundus image classification functionality ($\delta(p)=1$ when $p>0.5$, and $\delta(p)=0$ otherwise).
\end{itemize}
${p}(\mathbf{x}^r_l)$ is then used to determine whether the reference fundus image $\mathbf{x}^r_l$ is qualified to veto unreliable predictions. If $|{p}(\mathbf{x}^r_l)-{y}^r_l|>\kappa_2$, its vote will be omitted, where $\kappa_2$ is a threshold used to select qualified reference fundus images {(step 6 in Algorithm \ref{alg.ovv})}. 
In the meantime, the pre-trained MTSN can also ``contrastively-predict'' the scalar
\begin{equation}
	{p}_{l,k}^{r\rightarrow t}(\mathbf{x}^r_l, \mathbf{x}^t_k)=|p(\mathbf{x}^r_l)-q(\mathbf{x}^r_l,\mathbf{x}^t_k)|
	\label{eq.con_prob}
\end{equation}
indicating the GON probability as well as the label 
\begin{equation}
	\tilde{y}^{r\rightarrow t}_{l,k}(\mathbf{x}^r_l, \mathbf{x}^t_k)=|\delta(p(\mathbf{x}^r_l))-\delta(q(\mathbf{x}^r_l,\mathbf{x}^t_k))|
	\label{eq.con_label}
\end{equation}
of $\mathbf{x}^t_k$ from $\mathbf{x}^r_l$ using its input similarity measurement functionality\footnote{In rare cases, $\tilde{y}^{r\rightarrow t}_{l,k}(\mathbf{x}^r_l,\mathbf{x}^t_k)$ might not be equivalent to $\delta({p}_{l,k}^{r\rightarrow t}(\mathbf{x}^r_l, \mathbf{x}^t_k))$.}. {To determine the reliability of $\tilde{y}^t_k$ and whether it can be used as the pseudo label of $\mathbf{x}^t_k$}, all the reference fundus images $\{ \mathbf{x}^r_1,\dots,\mathbf{x}^r_m  \}\in \mathscr{X}^r$ in the mini-batch are used to provide additional judgements. Each pair of contrastively-predicted scalar (indicating GON probability) and label form a vote $({p}_{l,k}^{r\rightarrow t}(\mathbf{x}^r_l, \mathbf{x}^t_k),\tilde{y}^{r\rightarrow t}_{l,k}(\mathbf{x}^r_l, \mathbf{x}^t_k))$. 
With all votes collected from the qualified reference fundus images, the OVV self-training algorithm determines whether $\tilde{y}^t_k$ should be used as the pseudo label for $\mathbf{x}^t_k$
based on the following criteria ({step 9 in Algorithm \ref{alg.ovv}}):
\begin{itemize}
	\item Identical to the process of determining qualified reference fundus images, if any ${p}_{l,k}^{r\rightarrow t}(\mathbf{x}^r_l, \mathbf{x}^t_k)$ ($l\in[1,m]\cap\mathbb{Z}$) or ${p}(\mathbf{x}_k^t)$ is not close to 
	either 0 (healthy) or 1 (GON), as evaluated by the threshold $\kappa_2$, $\tilde{y}^t_k$ will not be assigned to $\mathbf{x}^t_k$.
	\item If a minority of more than $\kappa_1$ qualified reference fundus images disagree with the majority of the qualified reference fundus images, $\tilde{y}^t_k$ will not be assigned to $\mathbf{x}^t_k$. 
\end{itemize}
As discussed in Sect. \ref{sec.exp}, $\kappa_1=0$ ({all qualified reference images vote for the same category}) achieves the best overall performance. Therefore, the aforementioned  strategy is named ``\uline{\textit{One-Vote Veto}}'' in this paper. Since each target fundus image is required to be compared with all the reference fundus images in the same mini-batch, the proposed self-training strategy has a computational complexity of $\mathscr{O}(n^2)$, which is relatively memory-consuming. The reliable target fundus images and their pseudo labels are then included into the low-shot training data to fine-tune the pre-trained MTSN with supervised learning by minimizing a CWCE loss. The OVV self-training performance with respect to 
different $\kappa_1$, $\kappa_2$, and $m$ values is discussed in Sect. \ref{sec.exp}. 

\begin{table*}[!t]
	\fontsize{8.2}{10}\selectfont
	\caption{Comparison between supervised and low-shot learning (both on the low-shot OHTS training set). $\lambda$ is set to $0.3$. The best results are shown in bold font. }  
	\label{tab.lsl}
	\begin{center}
		\begin{tabular}{r|c|c|ccc|ccc}
			\toprule
			\multirow{2}{*}{Test set}
			
			&\multirow{2}{*}{Experiments}&\multirow{2}{*}{Training strategy}&\multicolumn{3}{c|}{\resnet-50}&\multicolumn{3}{c}{\mobilenet-v2}  \\
			
			\cline{4-9}
			&  & & Accuracy (\%) & F1-score (\%) & AUROC & Accuracy (\%) & F1-score (\%) & AUROC \\
			
			\hline
			\hline

			\multirow{3}{*}{ACRIMA \cite{acrima}} & Baseline & Supervised learning & 57.163 & 56.734 & 0.625 & \textbf{73.333} & \textbf{78.341} & 0.779\\
			
			& EWAD & Low-shot learning & \textbf{67.092} & \textbf{63.175}  & \textbf{0.758} & {70.355} & {67.797}  & 0.820 \\
			& EWSD & Low-shot learning &	49.504 & 25.523 & 0.437 & 66.241 & 59.107  & \textbf{0.823}\\
			
			\cline{1-9}
			
			\multirow{3}{*}{LAG \cite{lag}} & Baseline & Supervised learning & 64.318 & 57.445 & 0.714 & {65.122} & {63.268} & 0.781 \\
			& EWAD & Low-shot learning & \textbf{79.028} & 65.908 & \textbf{0.841} & \textbf{79.007}& \textbf{69.482}  & 0.843 \\
			& EWSD & Low-shot learning & 78.039 & \textbf{68.179}  & 0.826 &	78.430 &	61.830 &	\textbf{0.846} \\
			
			\cline{1-9}

			\multirow{3}{*}{\makecell{DIGS/ADAGES \cite{digs}}} & Baseline & Supervised learning &  59.639 & 59.708& 0.648 & 61.478 & 66.128 & 0.669 \\

			& EWAD & Low-shot learning & \textbf{67.745} & \textbf{60.754}  & \textbf{0.743} & \textbf{69.176} &	\textbf{68.145}  & \textbf{0.748}\\
			& EWSD  & Low-shot learning & 65.736 & 55.722  & 0.700  & 68.120 &	64.028 & 0.740\\
			
			\bottomrule
		\end{tabular}
	\end{center}
\end{table*}

\begin{table*}[!t]
	\fontsize{8.2}{10}\selectfont
	\centering	
	\caption{Evaluation of our OVV self-training w.r.t. different $\kappa_1$, $\kappa_2$, and $m$. The best results are shown in bold font. $\uparrow$ indicates that semi-supervised learning outperforms low-shot learning. }   
	\label{tab.ssl1} 
	\centering
	\begin{tabular}{c|ccc|ccc|ccc}
		\toprule
		\multirow{2}{*}{Dataset}
		& 
		\multirow{2}{*}{$\kappa_1$}
		& 
		\multirow{2}{*}{$\kappa_2$}
		&
		\multirow{2}{*}{$m$} 
		& 
		\multicolumn{3}{c}{\resnet-50}
		&
		\multicolumn{3}{c}{\mobilenet-v2}
		\\
		\cline{5-10}
		& & & & Accuracy (\%) & F1-score (\%) & AUROC & Accuracy (\%) & F1-score (\%) & AUROC 
		\\	
		\hline\hline
		
		\multirow{8}{*}{OHTS \cite{ohts1,ohts2}} & 0 & 0.01 & 20 & 91.415 $\uparrow$ & {41.148} $\uparrow$ &  {0.898} $\uparrow$ & 90.609 $\uparrow$ & {36.960} $\uparrow$ &  {0.887} $\uparrow$ \\
		& 0 & 0.01 & 15 & 92.113 $\uparrow$ & 	41.316 $\uparrow$ &  \textbf{0.898} $\uparrow$ & 93.470 $\uparrow$ & 36.976 $\uparrow$ &  \textbf{0.893} $\uparrow$ \\
		& 0 & 0.01 & 10 & \textbf{94.199} $\uparrow$ &	\textbf{43.759} $\uparrow$ &  0.890 $\uparrow$ & \textbf{93.742} $\uparrow$ & \textbf{37.779} $\uparrow$ & 	0.863 $\uparrow$ \\
		\cline{2-10}
		& 0 & 0.10  & 20 & 90.516 $\uparrow$ & \textbf{38.139} $\uparrow$ &  \textbf{0.898} $\uparrow$ & 88.825 $\uparrow$ & 34.351 $\uparrow$ &  \textbf{0.878} $\uparrow$	 \\
		& 2 & 0.01  & 20 & 90.717 $\uparrow$ & 35.818 $\uparrow$ &  0.885 $\uparrow$ & \textbf{93.463} $\uparrow$ & \textbf{35.204} $\uparrow$  & {0.862} $\uparrow$	 \\
		& 2 & 0.10  & 20 & 92.656 $\uparrow$ & 32.017 $\uparrow$ &  0.851 $\downarrow$ & 90.772 $\uparrow$ & 32.616 $\uparrow$  & 0.858 $\uparrow$ \\
		& 4 & 0.01  & 20 & 92.610 $\uparrow$ & 29.668  $\downarrow$ &  0.854 $\downarrow$	& 92.268 $\uparrow$ & 31.852 $\uparrow$  & 0.854 $\uparrow$ \\		
		& 4 & 0.10  &  20 & \textbf{92.672} $\uparrow$ & 28.463 $\downarrow$ &  0.842 $\downarrow$ & 90.803 $\uparrow$ & 32.690 $\uparrow$  & 0.859 $\uparrow$ \\
		\hline
		\multirow{3}{*}{ACRIMA \cite{acrima}} & 0 & 0.01 & 20 & {59.858} $\downarrow$ & {49.192} $\downarrow$ &  \textbf{0.775} $\uparrow$ & \textbf{72.340} $\uparrow$ & \textbf{70.229} $\uparrow$  & \textbf{0.840} $\uparrow$	 \\
		& 0 & 0.01 & 15 & \textbf{60.426} $\downarrow$ &	\textbf{49.365} $\downarrow$ & {0.751} $\downarrow$ & 63.404 $\downarrow$ & 54.895 $\downarrow$ &  {0.814} $\downarrow$ \\
		& 0 & 0.01 & 10 & {54.610} $\downarrow$ & {35.743} $\downarrow$ &  0.721 $\downarrow$ & 61.986 $\downarrow$ & 51.273 $\downarrow$ &  {0.826} $\uparrow$ \\
		\hline
		\multirow{3}{*}{LAG \cite{lag}} & 0 & 0.01 & 20 & \textbf{80.882} $\uparrow$ &	\textbf{66.304} $\uparrow$ &  \textbf{0.881} $\uparrow$ & \textbf{79.625} $\uparrow$ & \textbf{65.262} $\downarrow$ &  0.851 $\uparrow$  \\
		& 0 & 0.01 & 15 & {76.864} $\downarrow$ & {56.252} $\downarrow$ &  {0.825} $\downarrow$ & 76.670 $\downarrow$ & 55.906 $\downarrow$  & 0.841 $\downarrow$ \\
		& 0 & 0.01 & 10 & 	{76.638} $\downarrow$ & {56.518} $\downarrow$ &  {0.826} $\downarrow$ & 75.834 $\downarrow$ & 51.748 $\downarrow$  & \textbf{0.866} $\uparrow$ \\
		\hline
		\multirow{3}{*}{DIGS/ADAGES \cite{digs}} & 0 & 0.01 & 20 & \textbf{67.813} $\uparrow$ & \textbf{58.315} $\downarrow$ &  \textbf{0.763} $\uparrow$ & \textbf{69.653} $\uparrow$ & \textbf{63.258} $\downarrow$  & 0.777	$\uparrow$\\
		& 0 & 0.01 & 15 & {63.045} $\downarrow$ &	{44.727} $\downarrow$ &  {0.753} $\uparrow$ & 	66.383 $\downarrow$ & 52.888 $\downarrow$  & 0.773 $\uparrow$\\
		& 0 & 0.01 & 10 & 61.819 $\downarrow$ &	{41.340} $\downarrow$ &  0.727 $\downarrow$ & 	63.965 $\downarrow$ & 44.198 $\downarrow$  & \textbf{0.789} $\uparrow$\\
		\bottomrule
	\end{tabular}         
\end{table*}

\begin{sidewaystable}
	\fontsize{8.2}{10}\selectfont
	\caption{AUROC (shown along with 95\% CI) and training time $t$ (min) per epoch of supervised, low-shot, and semi-supervised glaucoma diagnosis. }
	\label{tab.auc_backbones}
	\centering
	\begin{tabular}{l|l|c|ccccc}
		\toprule
		\multicolumn{1}{c}{\multirow{1}{*}{Backbone}} & \multicolumn{1}{c}{\multirow{1}{*}{Method}} & \multicolumn{1}{c}{{\thead{Training strategy}}} &
		\multicolumn{1}{c}{\multirow{1}{*}{OHTS \cite{ohts1,ohts2}}} & \multicolumn{1}{c}{\multirow{1}{*}{ACRIMA \cite{acrima}}} & \multicolumn{1}{c}{\multirow{1}{*}{LAG \cite{lag}}} & \multicolumn{1}{c}{\multirow{1}{*}{DIGS/ADAGES \cite{digs}}} & 
		\multicolumn{1}{c}{\multirow{1}{*}{$t$ (min)}}
		\\  
		\hline
		\hline
		
		\multirow{3}{*}{\resnet-50} & Baseline & Supervised learning & {0.904 (95\% CI: 0.865, 0.935)} & 0.736 (95\% CI: 0.698, 0.771)  & 0.794 (95\% CI: 0.780, 0.807) & 0.744 (95\% CI: 0.696, 0.792) & 52.1
		\\ 
		
		& MTSN & Low-shot learning & {0.869} (95\% CI:  0.833, 0.901)& 0.758 (95\% CI:  0.723, 0.792) & {0.841} (95\% CI:  0.829, 0.853)  & {0.743} (95\% CI:  0.683, 0.795) & {1.8}
		\\ 
		& MTSN+OVV & Semi-supervised learning & 0.898 (95\% CI:  0.857, 0.928) & {0.775 (95\% CI:  0.741, 0.808)}& {0.881 (95\% CI:  0.870, 0.891)} & {0.763 (95\% CI:  0.695, 0.820)} & 203.7 
		\\ 
		\hline
		\multirow{3}{*}{\mobilenet-v2} 
		& Baseline & Supervised learning & {0.893 (95\% CI: 0.845, 0.932)} & 0.794 (95\% CI:  0.760, 0.825) &   {0.856 (95\% CI:  0.844, 0.867)} &   {0.786 (95\% CI:  0.728, 0.835)} & 42.9
		\\ 
		& MTSN  & Low-shot learning & {0.859} (95\% CI:  0.813, 0.896)& {0.820} (95\% CI:  0.786, 0.850)  & {0.843} (95\% CI:  0.831, 0.855) & {0.748} (95\% CI:  0.689, 0.802)  & {1.2}
		\\ 
		& MTSN+OVV & Semi-supervised learning & 0.887 (95\% CI:  0.850, 0.920) & {0.840 (95\% CI:  0.808, 0.867)} & 0.851 (95\% CI:  0.838, 0.862)  &  0.777 (95\% CI:  0.718, 0.826) & 125.4
		\\ 
		\hline
		\multirow{3}{*}{\densenet} 
		& Baseline & Supervised learning & 0.898 (95\% CI:  0.867, 0.927) & 0.810 (95\% CI:  0.778, 0.841) & 0.784 (95\% CI:  0.771, 0.798) & 0.743 (95\% CI:  0.688, 0.789) & 122.7
		\\ 
		& MTSN  & Low-shot learning & 0.854 (95\% CI:  0.811, 0.894) & 0.753 (95\% CI:  0.716, 0.786) & 0.853 (95\% CI:  0.842, 0.865) & 0.732 (95\% CI:  0.675, 0.785) & 5.7
		\\ 
		& MTSN+OVV & Semi-supervised learning & 0.896 (95\% CI:  0.861, 0.926) & 0.783 (95\% CI:  0.748, 0.817) & 0.831 (95\% CI:  0.818, 0.843) & 0.746 (95\% CI:  0.678, 0.800) & 324.0
		\\ 
		\hline
		\multirow{3}{*}{\efficientnet} 
		& Baseline & Supervised learning & 0.768 (95\% CI:  0.684, 0.834) & 0.633 (95\% CI:  0.590, 0.672) & 0.650 (95\% CI:  0.634, 0.667) & 0.658 (95\% CI:  0.611, 0.702) & 48.7
		\\ 
		& MTSN  & Low-shot learning & 0.863 (95\% CI:  0.818, 0.899) & 0.845 (95\% CI:  0.815, 0.873) & 0.845 (95\% CI:  0.833, 0.856) & 0.719 (95\% CI:  0.659, 0.774) & 1.5
		\\ 
		& MTSN+OVV & Semi-supervised learning & 0.886 (95\% CI:  0.845, 0.918) & 0.792 (95\% CI:  0.758, 0.824) & 0.850 (95\% CI:  0.837, 0.861) & 0.749 (95\% CI:  0.690, 0.800) & 159.8
		\\ 
		\bottomrule
	\end{tabular}
\end{sidewaystable}

\begin{table*}
	\fontsize{8.2}{10}\selectfont
	\caption{Comparisons of supervised learning, low-shot learning, and semi-supervised learning w.r.t. different percentages of labeled training data. In the baseline experiments (2.0\% of training data), one fundus photograph is selected from each patient; In the experiments with 0.5\% and 1.0\% of training data, a subset is created by choosing patient IDs at regular intervals. }
	\centering
	\begin{subtable}[h]{1\textwidth}
		\centering
		\begin{tabular}{c|ccc|ccc|ccc}
			\toprule
			\multirow{2}{*}{\makecell{Percentage of \\ training data}}  & \multicolumn{3}{c|}{Supervised learning} & \multicolumn{3}{c|}{Low-shot learning} & \multicolumn{3}{c}{Semi-supervised learning}\\
			
			\cline{2-10}
			
			& Accuracy (\%) & F1-score (\%) & AUROC & Accuracy (\%) & F1-score (\%) & AUROC & Accuracy (\%) & F1-score (\%) & AUROC \\
			
			\hline\hline
			
			0.5 & 80.402 & 18.066 & 0.720 & 84.769 & 20.228 & 0.759  & 88.856 & 25.351 & 0.797\\
			1.0 & 79.557 & 21.573 & 0.799 & 88.538 & 25.654 & 0.806 & 88.453 & 32.472 & 0.857\\
			\hline
			2.0 (baseline) & 84.141 & 26.743 & 0.838 & 87.150 & 31.726 & 0.865 & 91.415 & 41.148 & 0.898 \\
			\hline
			10.0 & 85.651 & 32.288 & 0.890 & 89.988 & 38.612 & 0.891 & 91.446 & 39.760 & 0.899\\
			50.0 & 92.950 & 40.188 & 0.907 & 94.223 & 40.826 & 0.887 & 92.067 & 38.187 & 0.889\\
			90.0 & 89.556 & 37.582 & 0.905 & 92.897 & 41.878 & 0.887 & 93.331 & 43.421 & 0.898
			\\
			
			\bottomrule
		\end{tabular}
		\caption{Backbone CNN: \resnet-50.}  
	\end{subtable}
	
	\centering
	\begin{subtable}[h]{1\textwidth}
		\centering
		\begin{tabular}{c|ccc|ccc|ccc}
			\toprule
			\multirow{2}{*}{\makecell{Percentage of \\ training data}}  & \multicolumn{3}{c|}{Supervised learning} & \multicolumn{3}{c|}{Low-shot learning} & \multicolumn{3}{c}{Semi-supervised learning}\\
			
			\cline{2-10}
			
			& Accuracy (\%) & F1-score (\%) & AUROC & Accuracy (\%) & F1-score (\%) & AUROC & Accuracy (\%) & F1-score (\%) & AUROC \\
			
			\hline\hline
			
			0.5 & 83.033 & 19.035 & 0.720 & 81.954 & 18.035 & 0.745 & 87.918 & 29.246	 & 0.826 \\
			1.0 & 67.659 & 16.458 & 0.748 & 91.764 & 20.628 & 0.772 & 89.407 & 31.357 & 0.841\\
			\hline
			2.0 (baseline) & 78.144 & 23.547 & 0.840 & 86.018 & 30.252 & 0.854 & 90.609 & 36.960 & 0.887 \\
			\hline
			10.0 & 90.429 & 34.873 & 0.884 & 89.562 & 36.808 & 0.897 & 89.624 & 36.887 & 0.888\\
			50.0 & 92.438 & 43.361 & 0.906 & 91.043 & 39.050 & 0.888 & 93.230 & 40.653 & 0.889 \\
			90.0 & 93.754 & 42.401 & 0.908 & 93.750 & 37.519 & 0.890 & 91.857 & 41.602 & 0.896
			\\
			\bottomrule
		\end{tabular}
		\caption{Backbone CNN: \mobilenet-v2.}  
	\end{subtable}
	\label{tab.percentage_exp}
\end{table*}

\section{Experiments}
\label{sec.exp}

\subsection{Datasets and Experimental Setups}
\label{sec.exp_datasets}
	
The datasets utilized in our experiments were collected by various clinicians in different countries using distinct fundus cameras. The ACRIMA \cite{acrima} and LAG \cite{lag} datasets are publicly available, while the OHTS \cite{ohts1, ohts2} and DIGS/ADAGES \cite{digs} datasets are available upon request after appropriate data use agreements are initiated. Their details are as follows:

\begin{itemize}
	\item
The \uline{\textbf{OHTS}} \cite{ohts1,ohts2} is the only multi-center longitudinal study that has precise information on the dates/timing of the development of glaucoma (the enrolled subjects did not have glaucoma at study entry) using standardized assessment criteria by an independent Optic Disc Reading Center and confirmed by three glaucoma specialist endpoint committee members.  In our experiments, a square region centered on the optic nerve head was first extracted from each raw fundus image using a well-trained DeepLabv3+ \cite{chen2017rethinking} model. A small part of the raw data are stereoscopic fundus images, each of which was split to produce two individual fundus images. Through this image pre-processing approach, a total of 74,678 fundus images were obtained. Moreover, \uline{ENPOAGDISC} (endpoint committee attributable to primary open angle glaucoma based on optic disc changes from photographs) \cite{fan2022detecting} labels are used as the classification ground truth. The fundus images are divided into a {training set (50,208 healthy images and 2,416 GON images)}, a validation set (7,188 healthy images and 426 GON images), and a test set (13,780 healthy images and 660 GON images) {by participant}. Splitting by participant (instead of by image) {ensured} that the validation and test sets did not contain images from any eyes or individuals used to train the model. More details on dataset preparation and baseline supervised learning experiments are provided in our recent publications \cite{fan2022detecting,fan2023transformer}. Additionally, we select one image (from only one eye) from each patient in the training set to create the low-shot training set (995 healthy images and 152 GON images). If both eyes of a patient do not convert to glaucoma in the study, the first captured fundus photograph is selected. If either eye of a patient converts to glaucoma in the study, the first glaucoma fundus photograph is selected. 

\item The \uline{\textbf{ACRIMA}} \cite{acrima} dataset consists of 309 healthy images and 396 GON images. It was collected as part of an initiative by the government of Spain. Classification was based on the review by a single experienced glaucoma expert. Images were excluded if they did not provide a clear view of the optic nerve head region \cite{christopher2020effects}. 

\item The \uline{\textbf{LAG}} \cite{lag} dataset contains 3,143 healthy images and 1,711 GON images\footnote{{The number of fundus images being published is fewer than what was reported in publication \cite{lag}. }}, collected by Beijing Tongren Hospital. Similar to the OHTS dataset, we also use the well-trained DeepLabv3+ \cite{chen2017rethinking} model to extract a square region centered on the optic nerve head from each fundus image. 

\item The \uline{\textbf{DIGS} and \textbf{ADAGES}} \cite{digs} are longitudinal studies designed to detect and monitor glaucoma based on optical imaging and visual function testing that, when combined, have generated tens of thousands of test results from over 4,000 healthy, glaucoma suspect, or glaucoma eyes. In our experiments, we utilize the DIGS/ADAGES test set (5,184 healthy images and 4,289 GON images) to evaluate the generalizability of our proposed methods. 
\end{itemize}

\begin{figure}[!t]
	\centering
	\begin{subfigure}{0.157\textwidth}
		\includegraphics[width=1.0\textwidth]{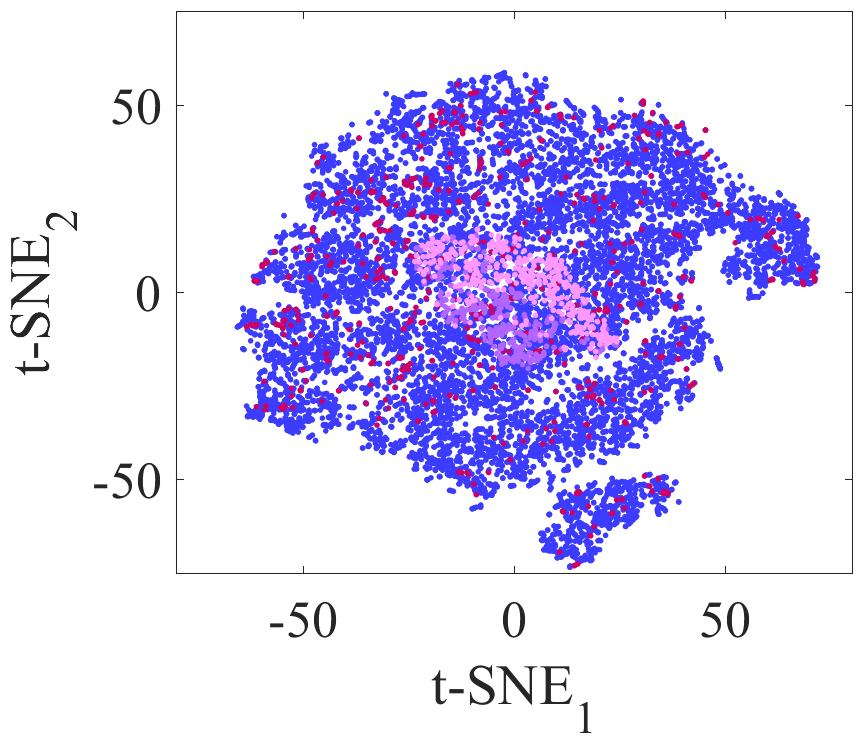}
		\caption{}
	\end{subfigure}
	\centering
	\begin{subfigure}{0.157\textwidth}
		\includegraphics[width=1.0\textwidth]{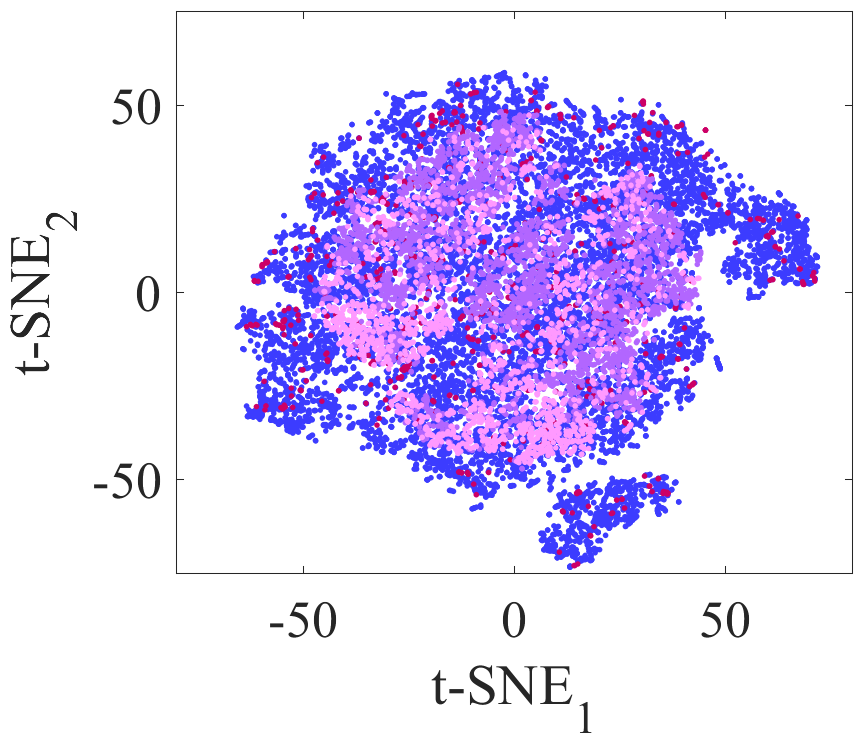}
		\caption{}
	\end{subfigure}
	\centering
	\begin{subfigure}{0.157\textwidth}
		\includegraphics[width=1.0\textwidth]{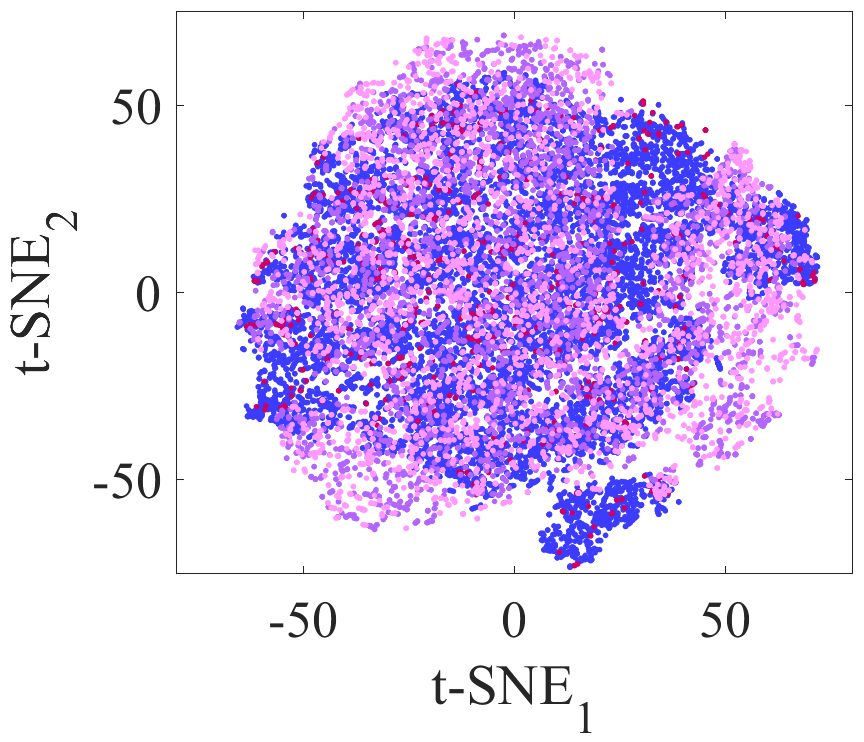}
		\caption{}
	\end{subfigure}
	\caption{
		Comparisons of dataset visualizations produced by t-SNE, where \textcolor{color1}{$\bullet$} and \textcolor{color2}{$\bullet$} represent the healthy and GON images in the OHTS test set, respectively; \textcolor{color3}{$\bullet$} and \textcolor{color4}{$\bullet$} in (a) represent the healthy and GON images in the ACRIMA dataset, respectively; \textcolor{color3}{$\bullet$} and \textcolor{color4}{$\bullet$} in (b) represent the healthy and GON images in the LAG  dataset, respectively; \textcolor{color3}{$\bullet$} and \textcolor{color4}{$\bullet$} in (c) represent the healthy and GON images in the DIGS/ADAGES test set, respectively.
	}
	\label{fig.tsne}
\end{figure}

Visualizations of the four test sets using t-SNE \cite{tsne} are provided in Fig. \ref{fig.tsne}. Since healthy and GON images are distributed similarly between the OHTS and LAG datasets, we expect models to perform similarly on these datasets. Dissimilar distributions in the ACRIMA and DIGS/ADAGES datasets led us to believe the performance of models on these datasets would be somewhat worse. Using these four datasets, we conduct three experiments:
\begin{enumerate}
	\item \textbf{Supervised learning experiment}: We employ transfer learning \cite{tan2018survey} to train \resnet-50 \cite{resnet}, \mobilenet-v2 \cite{mobilenetv2}, \densenet \cite{densenet}, and \efficientnet \cite{efficientnet} (pre-trained on the ImageNet database \cite{imagenet_cvpr09}), on the entire OHTS training set (including $\sim$53K fundus images). The best-performing models are selected using the OHTS validation set. Their performance is subsequently evaluated on the OHTS test set, the ACRIMA dataset, the LAG dataset, and the DIGS/ADAGES test set. 

	\item \textbf{Low-shot learning experiment}: The four pre-trained models mentioned above are used as the MTSN backbones and trained on the OHTS low-shot training set (containing 1,147 images) to validate the effectiveness of our proposed low-shot glaucoma diagnosis algorithm. The validation and testing procedures are identical to those in the supervised learning experiment. 
	
	\item \textbf{Semi-supervised learning experiment}: The MTSNs trained on the low-shot training set are fine-tuned on the entire OHTS training set without using additional ground-truth labels. The fine-tuned MTSNs are referred to as  MTSN+OVV. The validation and testing procedures are identical to those in the supervised learning experiment. 
\end{enumerate}

The fundus images are resized to $224\times224$ pixels. The initial learning rate is set to 0.001, which decays gradually after the 100th epoch. Due to the dataset imbalance problem, F1-score is utilized to select the best-performing models during the validation stage. Moreover, we adopt an early stopping mechanism during the validation stage to reduce over-fitting (the training will be terminated if the achieved F1-score has not increased for 10 epochs). We use three metrics: (1) accuracy, (2) F1-score, and (3) AUROC to quantify the performances of the trained models. While accuracy is generally reported in image classification papers, F1-score and AUROC are more comprehensive and reasonable evaluation metrics when the dataset is severely imbalanced. 

\subsection{Ablation study and Threshold Selection}
\label{sec.exp_fsl}

\begin{figure}[!t]
	\begin{subfigure}{0.49\textwidth}
		\centering
		\includegraphics[width=0.99\textwidth]{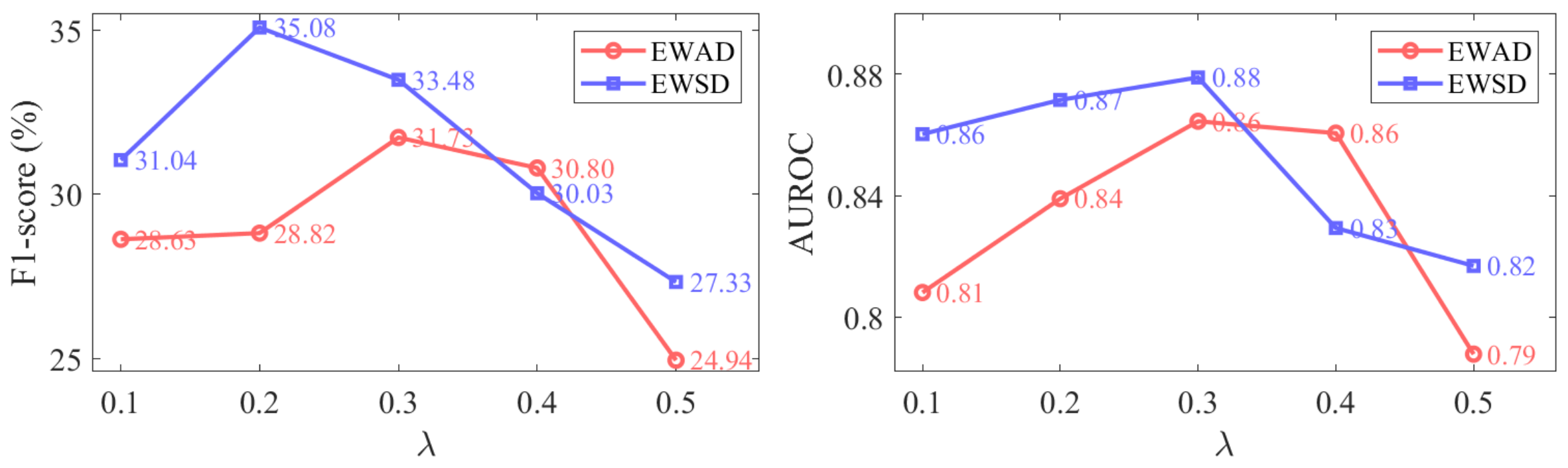}
		\caption{Backbone CNN: \resnet-50.  }
		\label{fig.lambda_resnet}
	\end{subfigure}
	\begin{subfigure}{0.49\textwidth}
		\centering
		\includegraphics[width=0.99\textwidth]{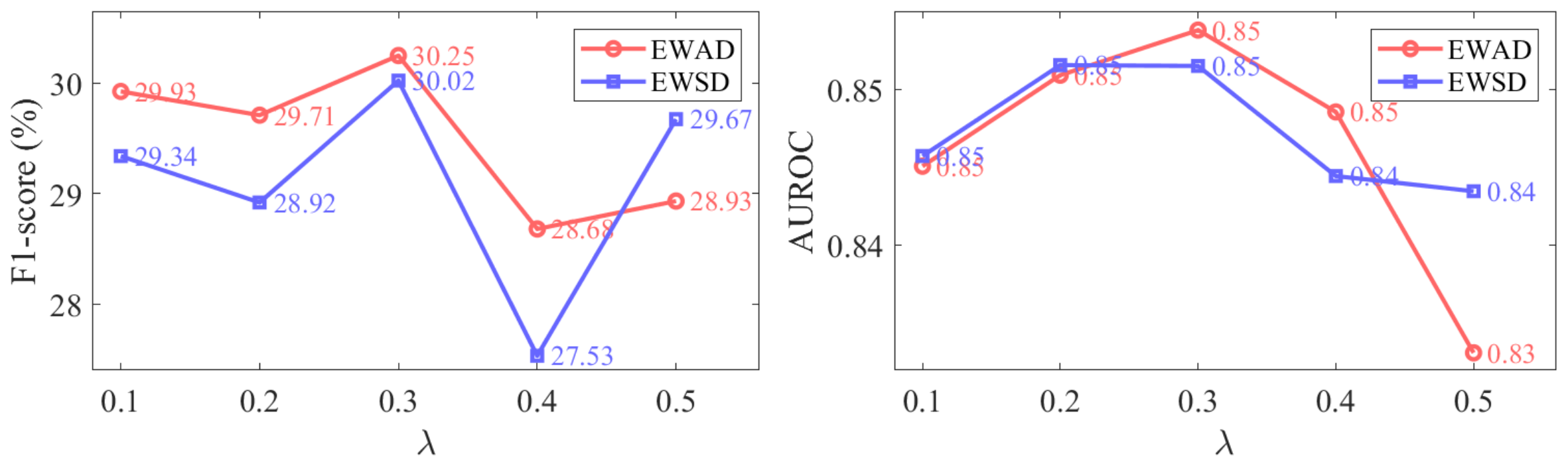}
		\caption{Backbone CNN: \mobilenet-v2.  }
		\label{fig.lambda_mobilenet}
	\end{subfigure}
	\caption{MTSN performance on the OHTS test set with respect to different $\lambda$ and $\Phi(\cdot)$.}
	\label{fig.MTSN_performance_wrt_lambda_phi}
\end{figure}

\begin{table*}[t!]
	\fontsize{8.2}{10}\selectfont
	\caption{Comparisons with other SoTA glaucoma diagnosis algorithms. The best results for each training strategy are shown in bold font.}  
	\label{tab.nine_additional_com}	
	
	\begin{subtable}[t!]{1\textwidth}
		\centering	
		\begin{tabular}{c|r|ccc}
			\toprule
			\multicolumn{1}{c}{Training strategy} & \multicolumn{1}{c}{Method} & Accuracy (\%) & F1-score (\%) & AUROC\\
			\hline
			\hline
			\multirow{6}{*}{\makecell{Supervised learning}} 
			& Li \etal \cite{li2018efficacy}  & 93.812	& 42.590  & 0.886  \\
			& G{\'o}mez-Valverde \etal \cite{gomez2019automatic}  & \textbf{95.068} &	\textbf{48.039} & 0.903 \\
			& Judy \etal \cite{judy2020automated} & 94.188 & 42.835 & 0.908 \\
			& Serener and Serte \cite{serener2019transfer}  & 90.492 & 41.227  & \textbf{0.912}\\
			& Thakur \etal \cite{thakur2020predicting} & 94.145 & 44.115  & 0.896\\
			& Fan \etal \cite{fan2022detecting} (Baseline Result) & 93.261 & 43.016  & 0.904\\
			
			\cline{1-5}
			
			\multirow{3}{*}{\makecell{Low-shot learning}} 
			& Kim \etal \cite{kim2017few} & 84.203	& 23.161 & 0.786 \\
			& \textbf{MTSN (Backbone: \resnet-50) (Ours)} & \textbf{87.150} & \textbf{31.726} & \textbf{0.865} \\
			& \textbf{MTSN (Backbone: \mobilenet-v2) (Ours)} & 86.018 & 30.252 & 0.854 \\
			
			\cline{1-5}
			
			\multirow{4}{*}{\makecell{Semi-supervised learning}} & Al Ghamdi \etal \cite{al2019semi} & 84.808 &	27.579 & 0.830	\\
			& Diaz-Pinto \etal \cite{diaz2019retinal} & 76.619 & 20.721 & 0.748 \\
			& \textbf{MTSN (Backbone: \resnet-50) + OVV Self-Training (Ours)} & \textbf{90.244} & \textbf{38.454} & \textbf{0.899}  \\
			& \textbf{MTSN (Backbone: \mobilenet-v2) + OVV Self-Training (Ours)}  & 89.360 & 36.599 & 0.891 \\
			
			\bottomrule
		\end{tabular}
		\caption{OHTS \cite{ohts2} test set. }
	\end{subtable}
	
	\begin{subtable}[h]{1\textwidth}
		\centering
		\begin{tabular}{c|r|ccc}
			\toprule
			\multicolumn{1}{c}{Training strategy} & \multicolumn{1}{c}{Method} & Accuracy (\%) & F1-score (\%) & AUROC\\
			\hline
			\hline
			\multirow{6}{*}{\makecell{Supervised learning}} 
			& Li \etal \cite{li2018efficacy}  & 60.142 & 46.272 & 0.813  \\
			& G{\'o}mez-Valverde \etal \cite{gomez2019automatic} & 63.546 & 53.358 & \textbf{0.826}  \\
			& Judy \etal \cite{judy2020automated} & 57.872 & 41.650 & 0.824 \\
			& Serener and Serte \cite{serener2019transfer} & 54.326 & 36.364 & 0.675 \\
			& Thakur \etal \cite{thakur2020predicting} & \textbf{65.106} & \textbf{58.020} & 0.794 \\
			& Fan \etal \cite{fan2022detecting} (Baseline Result) & 53.333 & 31.601 & 0.736 \\
			\cline{1-5}
			
			\multirow{3}{*}{\makecell{Low-shot learning}} 
			& Kim \etal \cite{kim2017few} & 64.965 & 58.627 & \textbf{0.844} \\
			& \textbf{MTSN (Backbone: \resnet-50) (Ours)} & 67.092 & 63.175 & 0.758 \\
			& \textbf{MTSN (Backbone: \mobilenet-v2) (Ours)} & \textbf{70.355} & \textbf{67.797} & 0.820\\
			
			\cline{1-5}
			
			\multirow{4}{*}{\makecell{Semi-supervised learning}} & Al Ghamdi \etal \cite{al2019semi} & 68.511 &	67.257 & 0.794 \\
			& Diaz-Pinto \etal \cite{diaz2019retinal} & 59.149 & 44.828 & 0.818 \\
			& \textbf{MTSN (Backbone: \resnet-50) + OVV Self-Training (Ours)} & 64.539 & 57.627 & 0.801 \\
			& \textbf{MTSN (Backbone: \mobilenet-v2) + OVV Self-Training (Ours)} & \textbf{72.340} & \textbf{70.939} & \textbf{0.835} \\
			
			\bottomrule
		\end{tabular}
		\caption{ACRIMA \cite{acrima} dataset. }

	\end{subtable}
	
	\begin{subtable}[h]{1\textwidth}
		\centering
		\begin{tabular}{c|r|ccc}
			\toprule
			\multicolumn{1}{c}{Training strategy} & \multicolumn{1}{c}{Method} & Accuracy (\%) & F1-score (\%) & AUROC\\
			\hline
			\hline
			\multirow{6}{*}{\makecell{Supervised learning}} 
A S			& Li \etal \cite{li2018efficacy}  & 76.535 & 53.756 & 0.855  \\
			& G{\'o}mez-Valverde \etal \cite{gomez2019automatic}  & 80.202 & 62.417 & \textbf{0.883} \\
			& Judy \etal \cite{judy2020automated} & 78.348 & 60.174 & 0.860 \\
			& Serener and Serte \cite{serener2019transfer}  & 77.379 & \textbf{66.545} & 0.806 \\
			& Thakur \etal \cite{thakur2020predicting} & \textbf{80.305} & 65.882 & 0.856 \\
			& Fan \etal \cite{fan2022detecting} (Baseline Result)  & 75.052 & 50.267 & 0.794 \\
			\cline{1-5}
			
			\multirow{3}{*}{\makecell{Low-shot learning}} 
			& Kim \etal \cite{kim2017few} & 74.619 & 65.000 & 0.805 \\
			& \textbf{MTSN (Backbone: \resnet-50) (Ours)} & \textbf{79.028} & 65.908 & 0.841 \\
			& \textbf{MTSN (Backbone: \mobilenet-v2) (Ours)} & 79.007 & \textbf{69.482} & \textbf{0.843} \\
			
			\cline{1-5}
			
			\multirow{4}{*}{\makecell{Semi-supervised learning}} 
			& Al Ghamdi \etal \cite{al2019semi} & 79.028 & \textbf{72.382} & 0.860	\\
			& Diaz-Pinto \etal \cite{diaz2019retinal} & 65.554 & 56.662	& 0.701 \\
			& \textbf{MTSN (Backbone: \resnet-50) + OVV Self-Training (Ours)} & \textbf{81.644} & 69.929 & \textbf{0.879} \\
			& \textbf{MTSN (Backbone: \mobilenet-v2) + OVV Self-Training (Ours)}  & 80.470 & 68.692 & 0.849 \\
			
			\bottomrule
		\end{tabular}
		\caption{LAG \cite{lag} dataset. }
	\end{subtable}

	\begin{subtable}[h]{1\textwidth}
		\centering
		\begin{tabular}{c|r|ccc}
			\toprule
			\multicolumn{1}{c}{Training strategy} & \multicolumn{1}{c}{Method} & Accuracy (\%) & F1-score (\%) & AUROC\\
			\hline
			\hline
			\multirow{6}{*}{\makecell{Supervised learning}}
			& Li \etal \cite{li2018efficacy}  & 65.293 & 46.955 & 0.780  \\
			& G{\'o}mez-Valverde \etal \cite{gomez2019automatic} & 65.157 & 45.729 & \textbf{0.795} \\
			& Judy \etal \cite{judy2020automated} & 63.556 & 41.144 & 0.760 \\
			& Serener and Serte \cite{serener2019transfer}  & 69.108 & 57.478 & 0.757 \\
			& Thakur \etal \cite{thakur2020predicting} & \textbf{70.606} & \textbf{60.322} & 0.786  \\
			& Fan \etal \cite{fan2022detecting} (Baseline Result)  & 62.500 & 38.042 & 0.744  \\
			
			\cline{1-5}
			
			\multirow{3}{*}{\makecell{Low-shot learning}} 
			& Kim \etal \cite{kim2017few} & 63.862	 & 52.949 & 0.687 \\
			& \textbf{MTSN (Backbone: \resnet-50) (Ours)} & 67.745 & 60.754 & 0.743 \\
			& \textbf{MTSN (Backbone: \mobilenet-v2) (Ours)} & \textbf{69.176} & \textbf{68.145} & \textbf{0.748} \\
			
			\cline{1-5}
			
			\multirow{4}{*}{\makecell{Semi-supervised learning}} & Al Ghamdi \etal \cite{al2019semi} & 64.850	& 54.816 & 0.716	\\
			& Diaz-Pinto \etal \cite{diaz2019retinal} & 64.441 & 59.908 & 0.677 \\
			& \textbf{MTSN (Backbone: \resnet-50) + OVV Self-Training (Ours)} & 66.281 & 55.486 & 0.747 \\
			& \textbf{MTSN (Backbone: \mobilenet-v2) + OVV Self-Training (Ours)} & \textbf{70.402} & \textbf{66.015} & \textbf{0.776} \\
			
			\bottomrule
		\end{tabular}
		\caption{DIGS/ADAGES \cite{digs} test set. }
	\end{subtable}
\end{table*}

We set $\lambda$ in (\ref{eq.L}) to 0.1, 0.2, 0.3, 0.4, and 0.5, respectively, and compare the MTSN performance when $\Phi$ computes the element-wise absolute difference (EWAD) and element-wise squared difference (EWSD), respectively. The comparisons in terms of F1-score and AUROC on the OHTS test set are provided in Fig. \ref{fig.MTSN_performance_wrt_lambda_phi}. It can be seen that the MTSN achieves the best overall performance when $\lambda=0.3$. This is reasonable, as a higher $\lambda$ weighs more on the image classification task, easily resulting in over-fitting. Additionally, the MTSN in which $\Phi(\cdot)$ computes the EWSD between $\mathbf{h}_i$ and $\mathbf{h}_j$ performs better when using \resnet-50 as the backbone CNN but slightly worse when using \mobilenet-v2 as the backbone CNN. Therefore, we further evaluate their generalizability on three additional test sets, as shown in Table \ref{tab.lsl}. When $\Phi(\cdot)$ computes the EWAD, the MTSN generally performs better or very similarly on the additional test sets, especially when testing the MTSN assembled with \resnet-50 on the ACRIMA dataset. EWAD is, therefore, used in the following experiments. Furthermore, Table \ref{tab.lsl} provides the results of a baseline supervised learning experiment conducted on the low-shot training set. The results suggest that low-shot learning performs much better than supervised learning when the training size is small.

Furthermore, we discuss the selection of the thresholds $\kappa_1$ and $\kappa_2$ (used to select reliable ``self-predictions'' and ``contrastive predictions'' in our OVV self-training) as well as the impact of different mini-batch sizes $2m$ on OVV self-training (each mini-batch contains $m$ pairs of reference and target fundus images). Table \ref{tab.ssl1} shows the MTSN performances with respect to different $\kappa_1$, $\kappa_2$, and $m$. When evaluated on the OHTS test set, it can be seen that accuracy and F1-score increase slightly, but AUROC almost remains the same, with the decrease of $m$. Moreover, with the increase of $\kappa_1$ and $\kappa_2$, the standard to determine reliable predictions becomes lower, making the semi-supervised learning performance degrade. Based on this experiment, we believe OVV self-training benefits from smaller $\kappa_1$ and $\kappa_2$. 

Additionally, MTSNs, trained under different $m$, are evaluated on the three additional test sets, as shown in Table \ref{tab.ssl1}. It can be seen that the network trained with a larger $m$ typically shows better results. When $m$ decreases, the generalizability of MTSN degrades dramatically, especially for F1-score (decreases by around 9-19\%). Therefore, increasing the mini-batch size can improve the MTSN generalizability, as more reference fundus images are used to provide contrastive predictions for the target fundus images, which can veto more unreliable predictions on the unlabeled data. {Hence, we increase $m$ to 30 to further improve OVV self-training when comparing it with other published SoTA algorithms, as shown in Sect. \ref{sec.com_SoTA}}. Since our threshold selection experiments cover a very limited number of discrete sets of $\kappa_1$, $\kappa_2$, and $m$, we believe better performance can be achieved when more values are tested.

\subsection{Comparison of Supervised, Low-Shot, and Semi-Supervised Glaucoma Diagnosis}
\label{sec.exp_supervised_learning}

Comparisons of supervised learning, low-shot learning, and semi-supervised learning (w.r.t. four backbone CNNs: \resnet-50, \mobilenet-v2, DenseNet, and EfficientNet) for glaucoma diagnosis are provided in Table \ref{tab.auc_backbones}. First, these results suggest that the MTSNs fine-tuned with OVV self-training that requires a small number of labeled fundus images perform similarly (AUROC  95\% CI overlaps considerably) and, in some cases, significantly better (AUROC  95\% CI does not overlap) than the backbone CNNs trained with a large number of labeled fundus images (50 times larger) under full supervision. 

Specifically, when using \resnet-50, \mobilenet-v2, or \densenet as the backbone CNN, semi-supervised learning performs similarly to supervised learning on the OHTS and DIGS/ADAGES test sets, and in most cases, significantly better than supervised learning on the ACRIMA and LAG datasets. Although \efficientnet trained through supervised learning performs unsatisfactorily on all four test sets, it shows considerable compatibility with MTSN in the low-shot and semi-supervised learning experiments. Second as expected, the AUROC scores achieved by low-shot learning are in most, but not all, cases slightly lower than those achieved by the backbone CNNs, when evaluated on the OHTS test set. However, low-shot learning shows better generalizability than supervised learning on the ACRIMA and LAG datasets. Moreover, since low-shot learning uses only a small amount of training data, training an MTSN is much faster than supervised learning. As MTSNs assembled with \resnet-50 and \mobilenet-v2 typically demonstrate better performances than the ones assembled with \densenet and EfficientNet, we only use the former two CNNs for the following experiments. 

\begin{figure*}[!t]
	\centering
	\begin{subfigure}{0.495\textwidth}
		\includegraphics[width=1.0\textwidth]{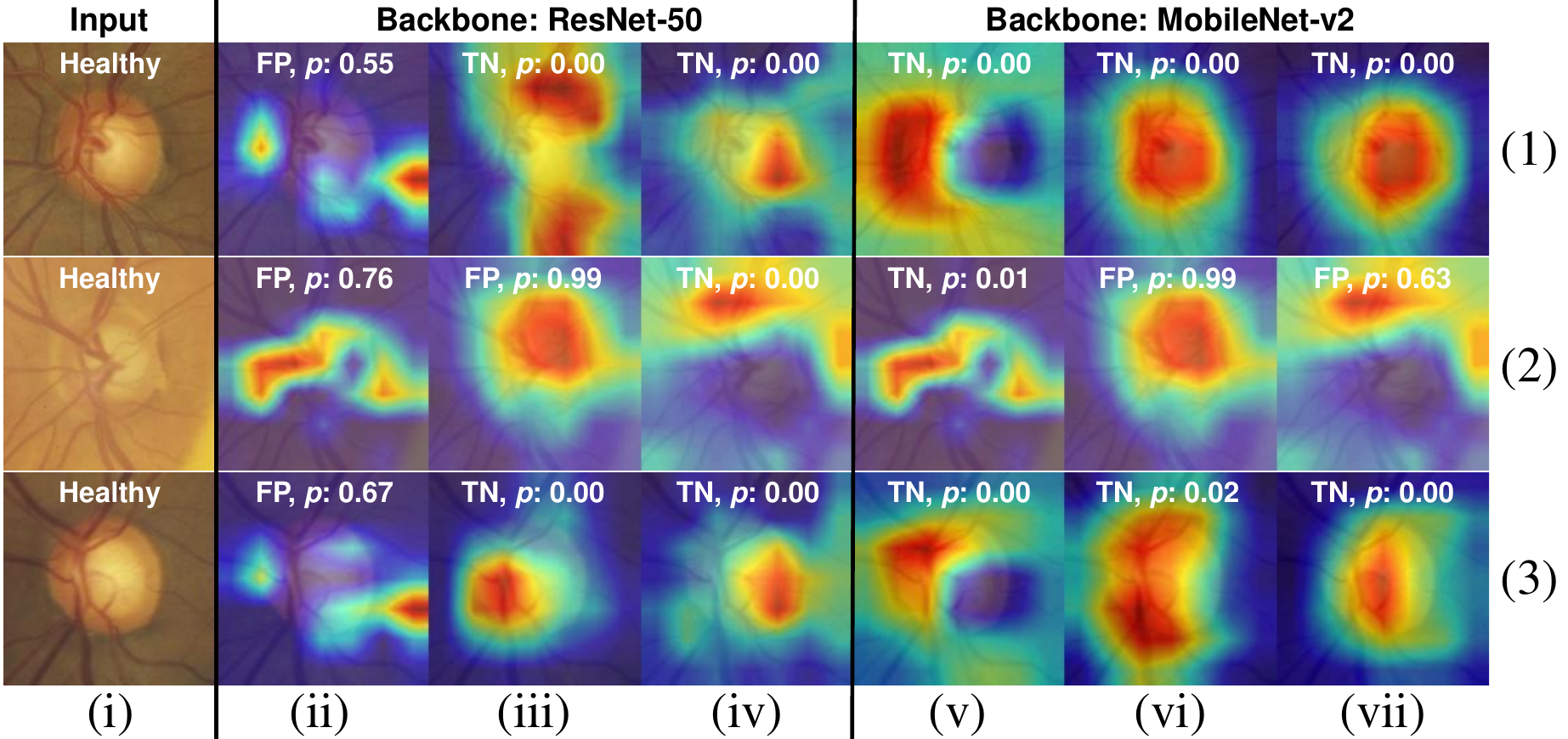}
		\caption{OHTS \cite{ohts3} healthy fundus images.}
	\end{subfigure}
	\centering
	\begin{subfigure}{0.495\textwidth}
		\includegraphics[width=1.0\textwidth]{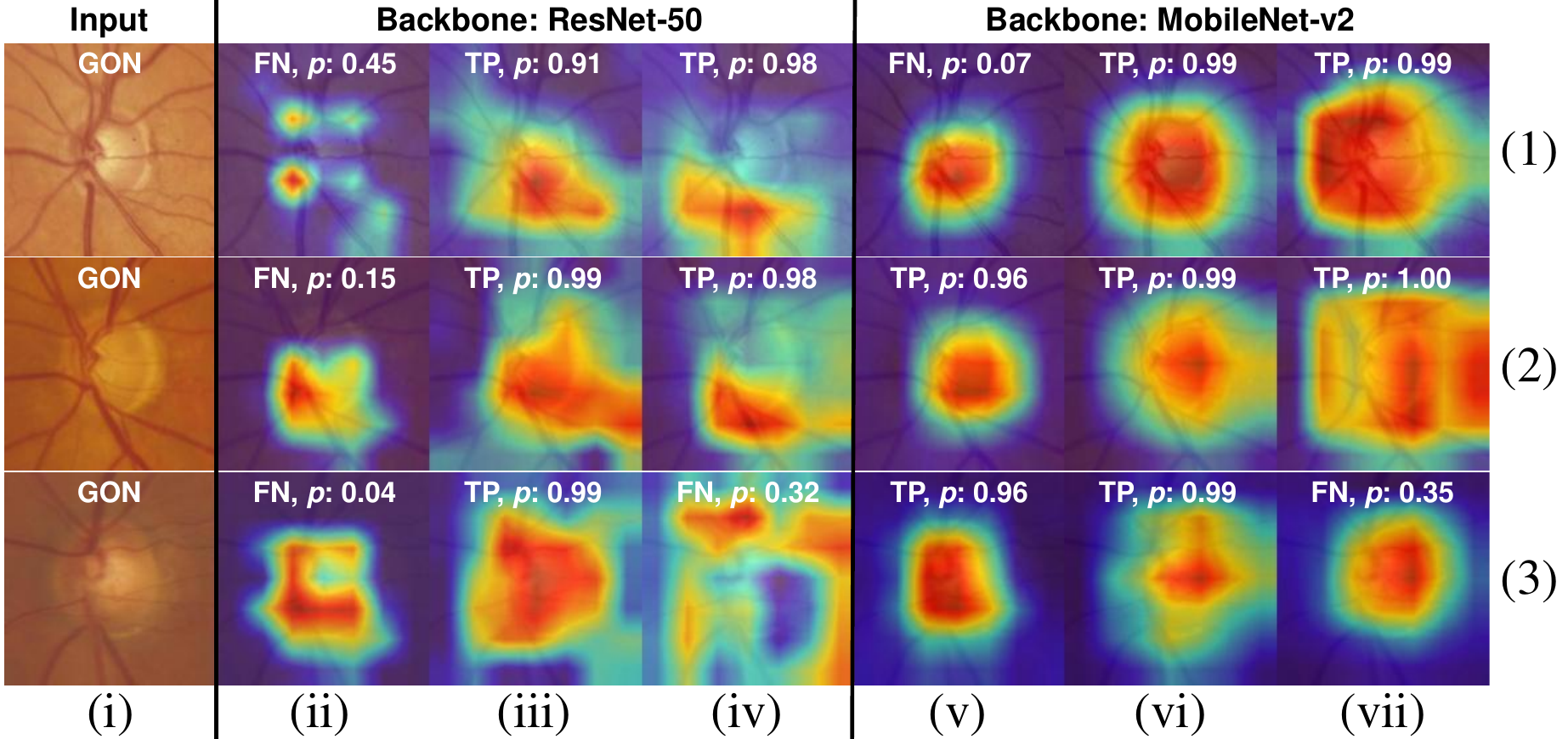}
		\caption{OHTS \cite{ohts3} GON fundus images.}
	\end{subfigure}
	\\
	\centering
	\begin{subfigure}{0.495\textwidth}
		\includegraphics[width=1.0\textwidth]{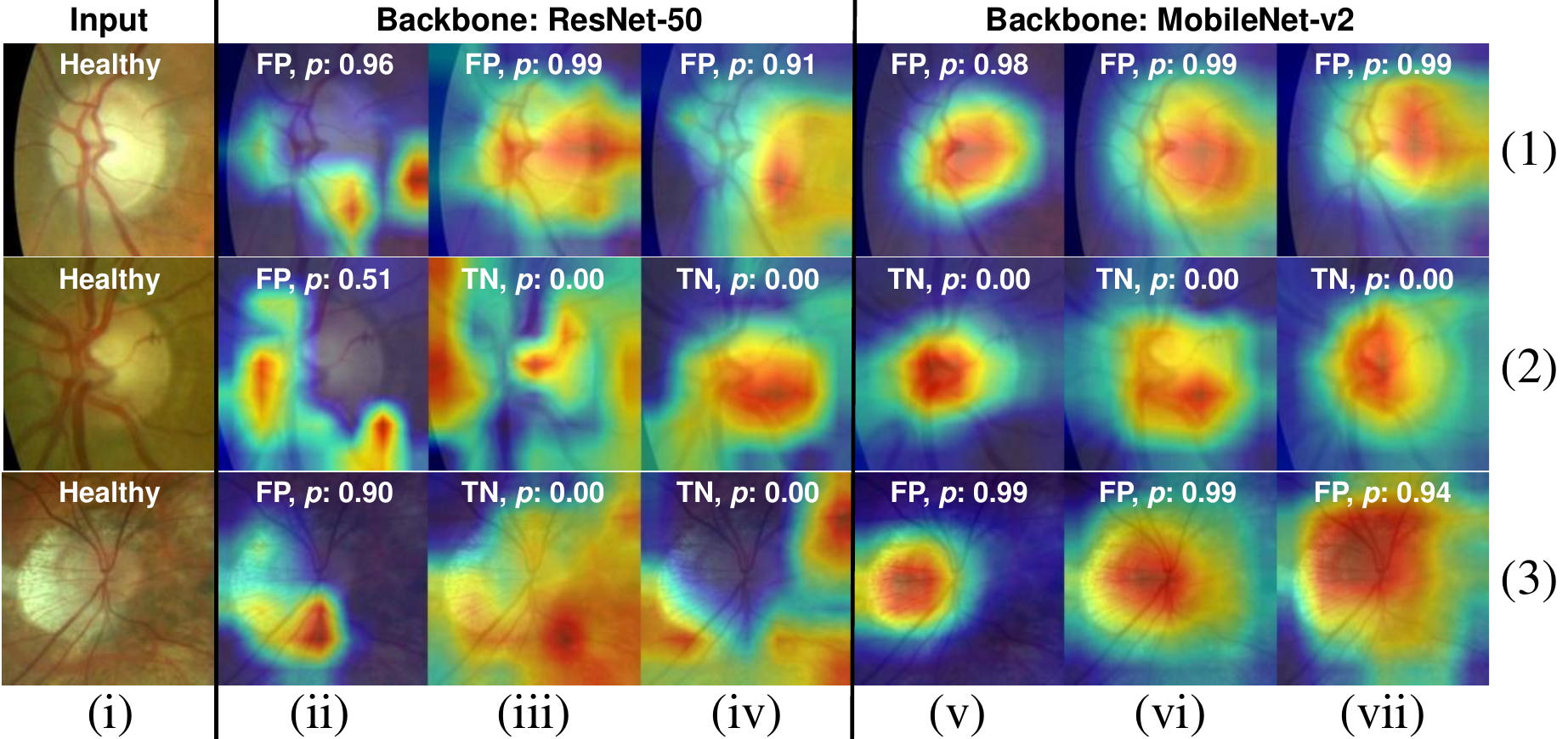}
		\caption{ACRIMA \cite{acrima} healthy fundus images.}
	\end{subfigure}
	\centering
	\begin{subfigure}{0.495\textwidth}
		\includegraphics[width=1.0\textwidth]{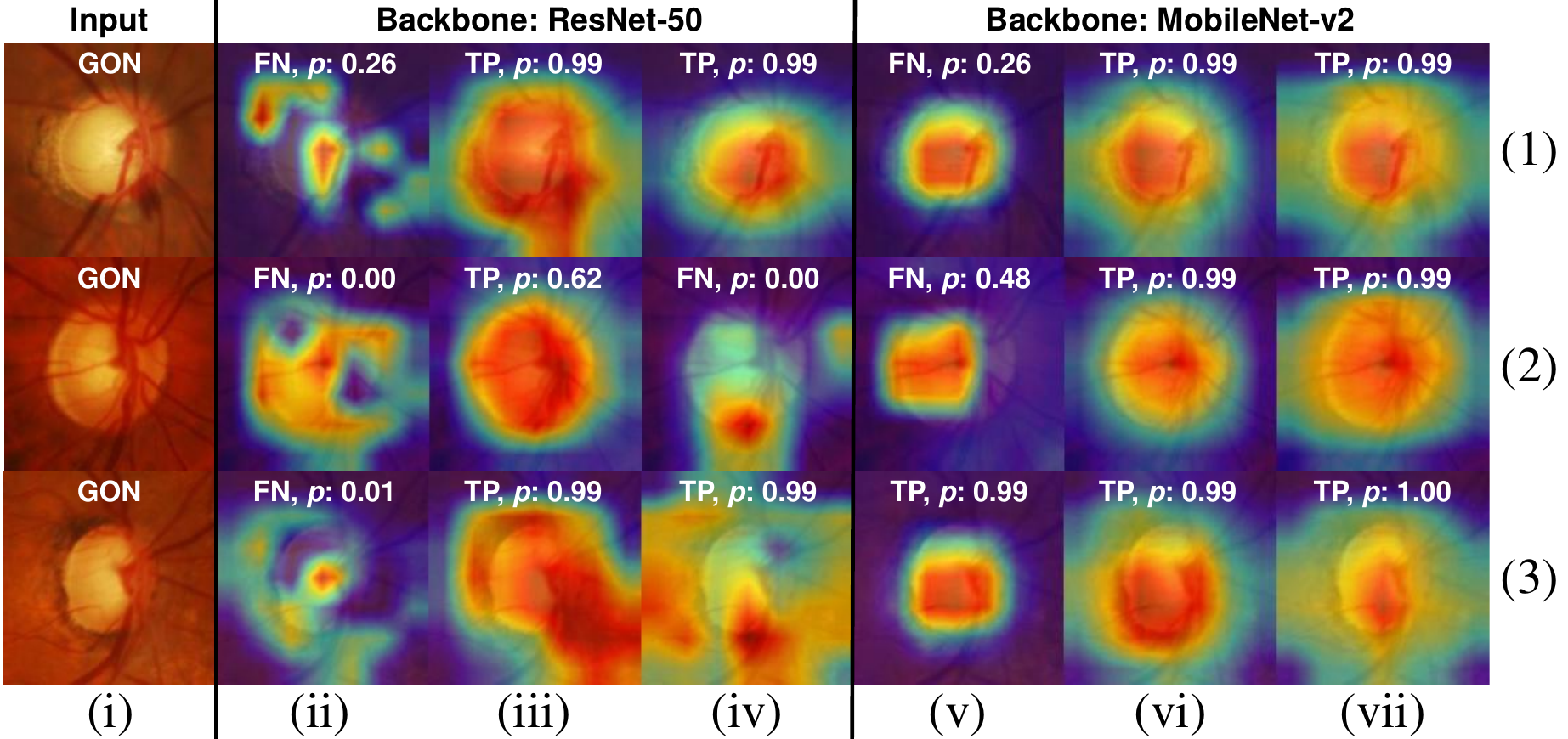}
		\caption{ACRIMA \cite{acrima} GON  fundus images.}
	\end{subfigure}
	\\
	\centering
	\begin{subfigure}{0.495\textwidth}
		\includegraphics[width=1.0\textwidth]{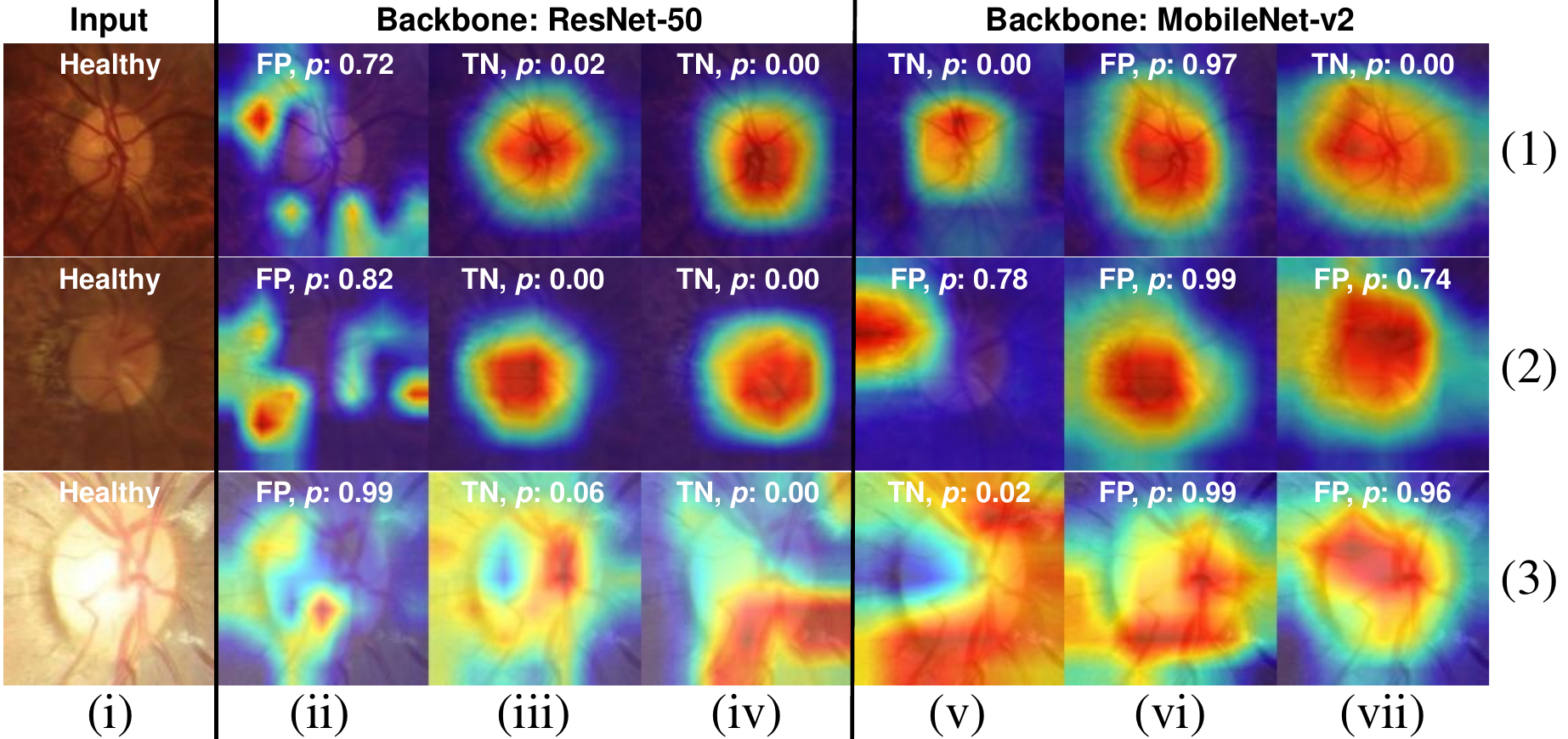}
		\caption{LAG \cite{lag} healthy fundus images.}
	\end{subfigure}
	\centering
	\begin{subfigure}{0.495\textwidth}
		\includegraphics[width=1.0\textwidth]{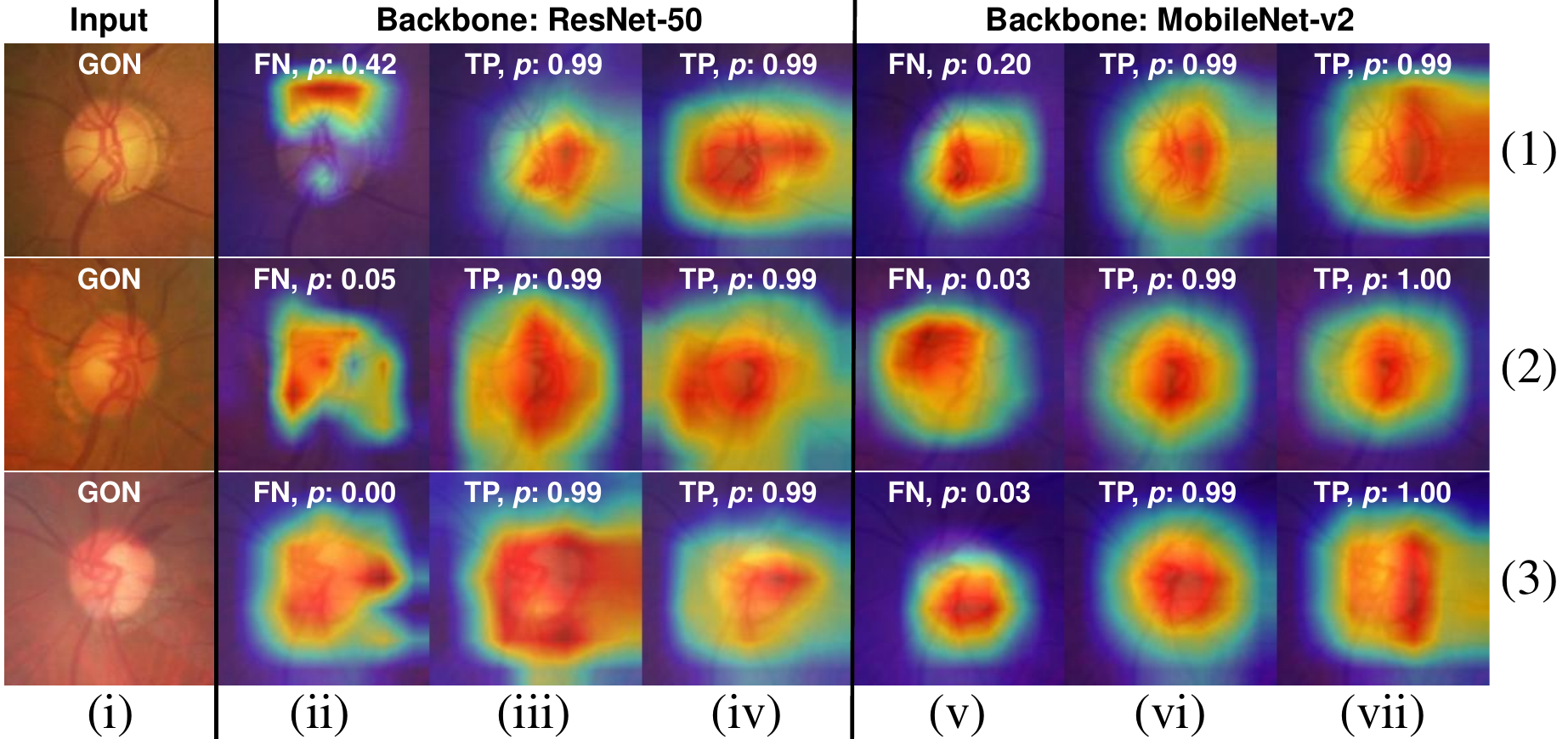}
		\caption{LAG \cite{lag} GON fundus images. }
	\end{subfigure}
	\\
	\centering
	\begin{subfigure}{0.495\textwidth}
		\includegraphics[width=1.0\textwidth]{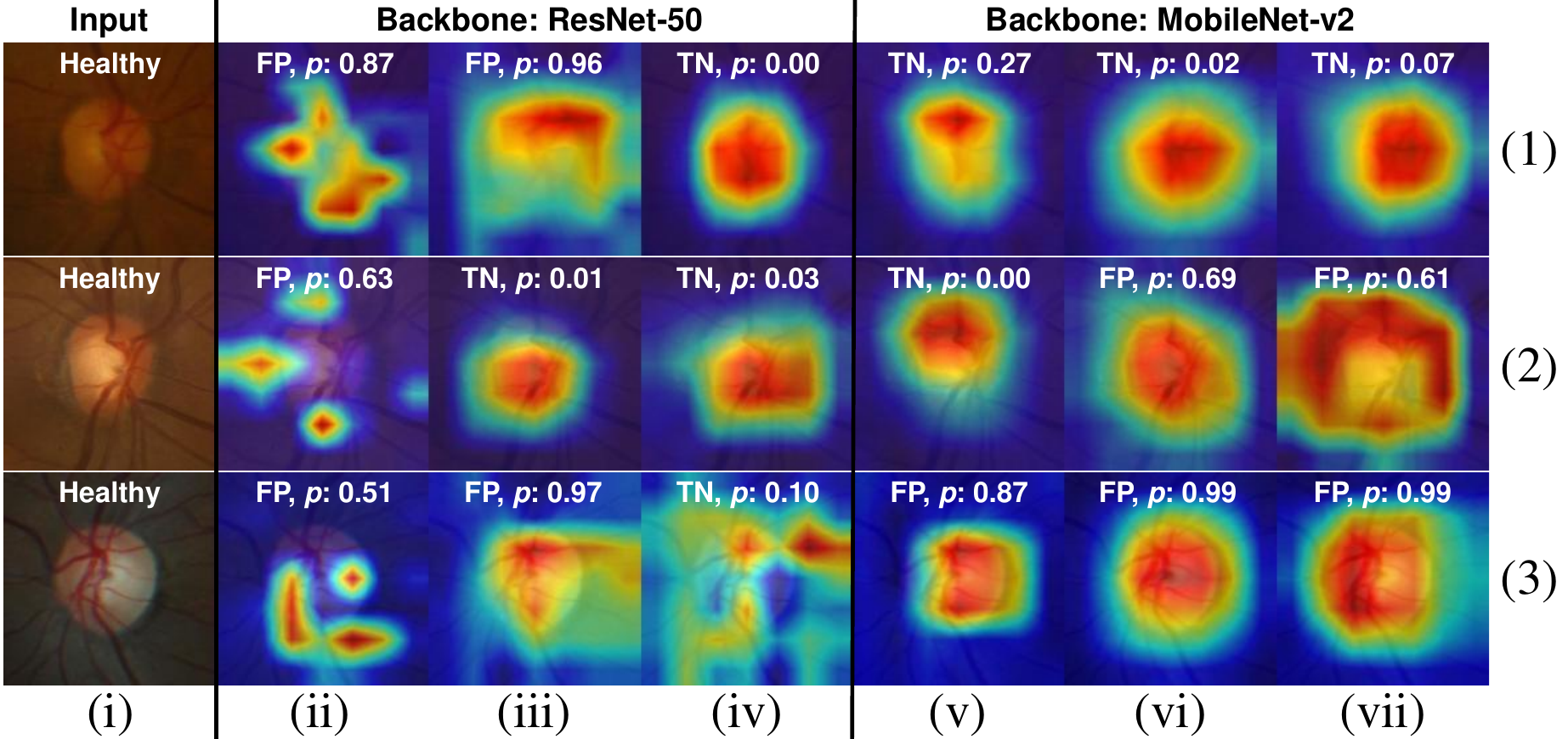}
		\caption{DIGS/ADAGES \cite{digs} healthy fundus images.}
	\end{subfigure}
	\centering
	\begin{subfigure}{0.495\textwidth}
		\includegraphics[width=1.0\textwidth]{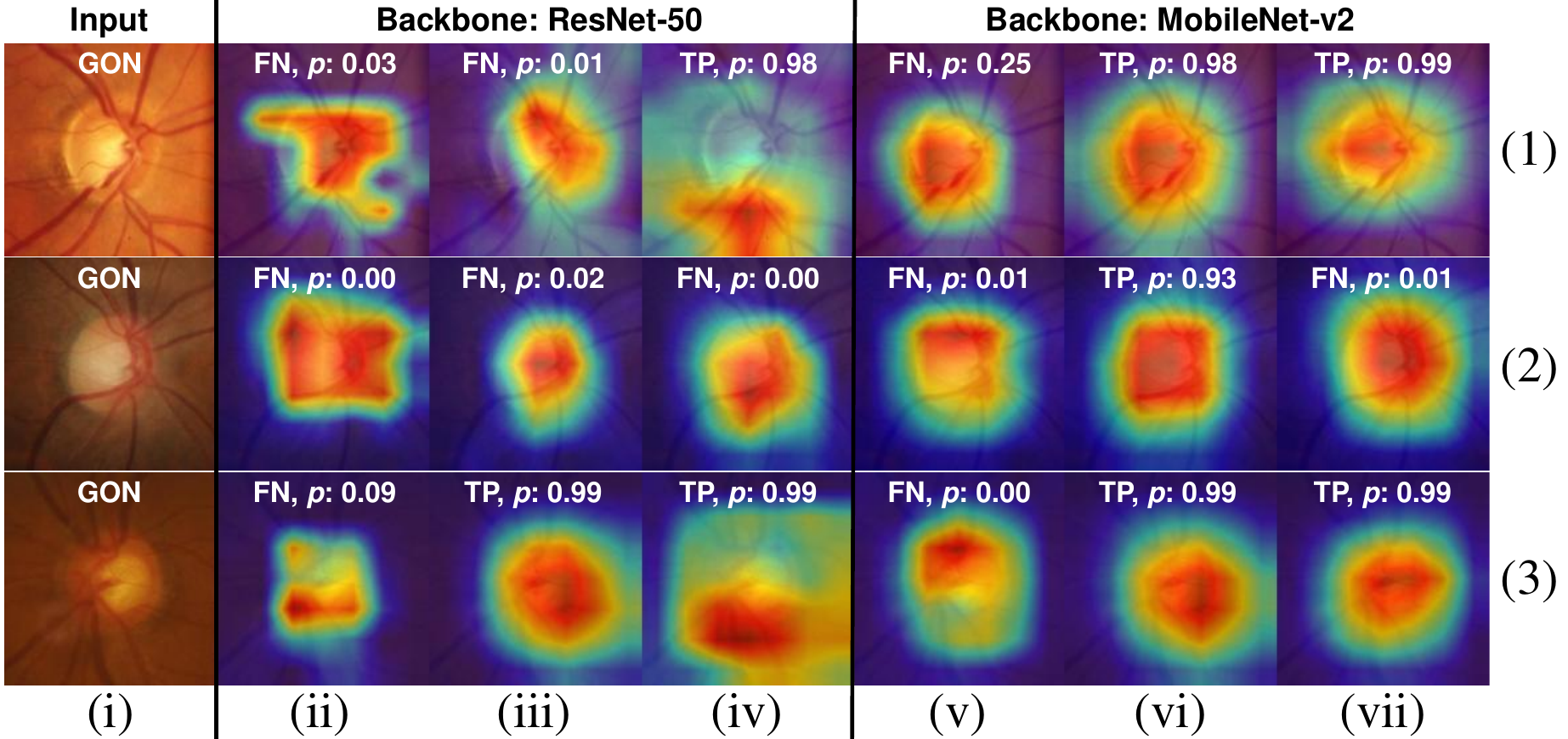}
		\caption{DIGS/ADAGES \cite{digs} GON fundus images.}
	\end{subfigure}
	\caption{
		Examples of Grad-CAM++ \cite{gradcamplus} results: (i) fundus images; (ii) and (v) show the class activation maps of (i), obtained by the backbone CNNs trained through supervised learning on the entire training set (containing $\sim$53K fundus images); (iii) and (vi) show the class activation maps of (i), obtained by MTSNs trained through low-shot learning on a small training set (containing 1,147 fundus images);
		(iv) and (vii) show the class activation maps of (i), obtained by MTSNs fine-tuned with our proposed OVV self-training on the entire training set (containing $\sim$53K fundus images without ground-truth labels).
	}
	\label{fig.resnet50_visualization}	
\end{figure*}

We also employ Grad-CAM++ \cite{gradcamplus} to explain the models' decision-making, as shown in Fig. \ref{fig.resnet50_visualization}.
These results suggest that the optic nerve head areas impact model decisions most. The neuroretinal rim areas are identified as most important, and the periphery contributed comparatively little to model decisions for both healthy and GON eyes \cite{christopher2018performance}. 

We also carry out a series of experiments with respect to different percentages of training data, as shown in Table \ref{tab.percentage_exp}, to further validate the effectiveness of our proposed low-shot and semi-supervised learning algorithms. The backbone CNNs trained via supervised learning on the small subsets generally perform worse than the MTSNs trained via low-shot and semi-supervised learning. With the labeled training data increase, the models' performance gets saturated. When using less labeled training data ($0.5\%$ and $1.0\%$), the MTSN performance degrades. However, its performance can still be greatly improved with OVV self-training (accuracy, F1-score, and AUROC can be improved by up to 6\%, 11\%, and 0.08, respectively). In addition, when using over $10\%$ of the entire training data, the MTSN performance saturates, and the OVV self-training can bring very limited improvements on MTSNs. 

\subsection{Comparisons with other SoTA glaucoma diagnosis approaches}
\label{sec.com_SoTA}
Table \ref{tab.nine_additional_com} provides comprehensive comparisons with nine SoTA glaucoma diagnosis algorithms\footnote{Our recent work \cite{fan2022detecting} provides the baseline supervised learning results.}. The results suggest that (a) for low-shot learning, the MTSNs trained by minimizing our proposed CWCE loss perform significantly better than the SoTA low-shot glaucoma diagnosis approach \cite{kim2017few} on all four datasets (accuracy, F1-score, and AUROC are up to 5\%, 15\%, and 0.08 higher, respectively), and (b) for semi-supervised learning, the MTSNs fine-tuned with OVV self-training also achieve the superior performances over another two SoTA semi-supervised glaucoma diagnosis approaches \cite{al2019semi,diaz2019retinal} (accuracy, F1-score, and AUROC are up to 6\%, 11\%, and 0.07 higher, respectively). Compared with the SoTA supervised approaches, the fine-tuned MTSNs demonstrate similar performance on the OHTS test set and better generalizability on three additional test sets. Therefore, we believe that MTSN with our proposed OVV self-training is an effective technique for semi-supervised glaucoma diagnosis.

\subsection{Comparisons with SoTA general-purpose semi-supervised learning approaches}
\label{sec.com_sota_semi}

	\begin{table*}[t!]
	\fontsize{8.2}{10}\selectfont
	\caption{Comparisons with SoTA general-purpose semi-supervised learning methods that use the vision Transformer \cite{dosovitskiyimage} as the backbone network. The best results of each dataset are shown in bold font.}  
	\label{tab.comparisons_sota_semi_supervised_methods}	
	
	\centering
	{
		\begin{tabular}{c|c|ccc}
			\toprule
			\multicolumn{1}{c}{Test Set} & \multicolumn{1}{c}{Method} & Accuracy (\%) & F1-score (\%) & AUROC\\
			\hline
			\hline
			\multirow{5}{*}{\makecell{OHTS \cite{ohts1,ohts2}}} 
			& FreeMatch \cite{wang2022freematch}  & 86.510 & 27.745 & 0.784  \\
			& SoftMatch \cite{chen2023softmatch}  & 85.679 & 25.718 & 0.768 \\
			& FixMatch \cite{sohn2020fixmatch}    & 84.695 & 27.825 & 0.807 \\
			& FlexMatch \cite{zhang2021flexmatch} & \textbf{92.715} & 26.331 & 0.696 \\
			& \textbf{OVV Self-Training (Ours)} & {90.244} & \textbf{38.454} & \textbf{0.899}  \\
			\cline{1-5}
			
			\multirow{5}{*}{\makecell{ACRIMA \cite{acrima}}} 
			& FreeMatch \cite{wang2022freematch}  & 77.872 & 77.778 & 0.791  \\
			& SoftMatch \cite{chen2023softmatch}  & 78.582 & 79.115 & {0.795} \\
			& FixMatch \cite{sohn2020fixmatch}    & \textbf{78.582} & \textbf{79.622} & 0.792 \\
			& FlexMatch \cite{zhang2021flexmatch} & 60.426 & 47.850 & 0.644 \\
			& \textbf{OVV Self-Training (Ours)} & 64.539 & 57.627 & \textbf{0.801}  \\
			\cline{1-5}
			\cline{1-5}
			
			\multirow{5}{*}{\makecell{LAG \cite{lag}}} 
			& FreeMatch \cite{wang2022freematch}  & 78.471 & 67.087 & 0.793  \\
			& SoftMatch \cite{chen2023softmatch}  & 79.110 & 67.972 & 0.798 \\
			& FixMatch \cite{sohn2020fixmatch}    & 81.314 & \textbf{73.394} & 0.827 \\
			& FlexMatch \cite{zhang2021flexmatch} & 71.302 & 35.360 & 0.654 \\
			& \textbf{OVV Self-Training (Ours)} & \textbf{81.644} & 69.929 & \textbf{0.879}  \\
			\cline{1-5}
			
			\cline{1-5}
			
			\multirow{5}{*}{\makecell{DIGS/ADAGES \cite{digs}}} 
			& FreeMatch \cite{wang2022freematch}  & 65.429 & 56.716 & 0.685  \\
			& SoftMatch \cite{chen2023softmatch}  & 65.395 & 56.986 & 0.679 \\
			& FixMatch \cite{sohn2020fixmatch}    & \textbf{68.835} & \textbf{65.090} & 0.731 \\
			& FlexMatch \cite{zhang2021flexmatch} & 57.050 & 26.558 & 0.594 \\
			& \textbf{OVV Self-Training (Ours)} & 66.281 & 55.486 & \textbf{0.747}  \\
			
			\bottomrule
		\end{tabular}
	}
\end{table*}
Table \ref{tab.comparisons_sota_semi_supervised_methods} provides a comprehensive comparison of our proposed OVV self-training approach with four SoTA general-purpose semi-supervised learning methods: FreeMatch \cite{wang2022freematch}, SoftMatch \cite{chen2023softmatch}, FixMatch \cite{sohn2020fixmatch}, and FlexMatch \cite{zhang2021flexmatch}, which all employ vision Transformer \cite{dosovitskiyimage} as their backbone network. Our results demonstrate that the proposed OVV self-training approach outperforms these methods in terms of F1-score and AUROC on the OHTS dataset. Specifically, we observe improvements in F1-score ranging from approximately 11\% to 13\%, and improvements in AUROC ranging from around 9\% to 20\%. Furthermore, our method demonstrates better generalizability in terms of AUROC across three additional fundus image test sets. Although their results are inferior to ours, particularly in terms of AUROC, it may be unjust to compare them as they were not specifically designed for the diagnosis of glaucoma or other diseases.

\subsection{MTSN and CWCE Loss for Few-Shot Multi-Class Biomedical Image Classification}
\label{sec.other_applications}

\begin{figure*}[!t]
	\centering
	\includegraphics[width=0.999\textwidth]{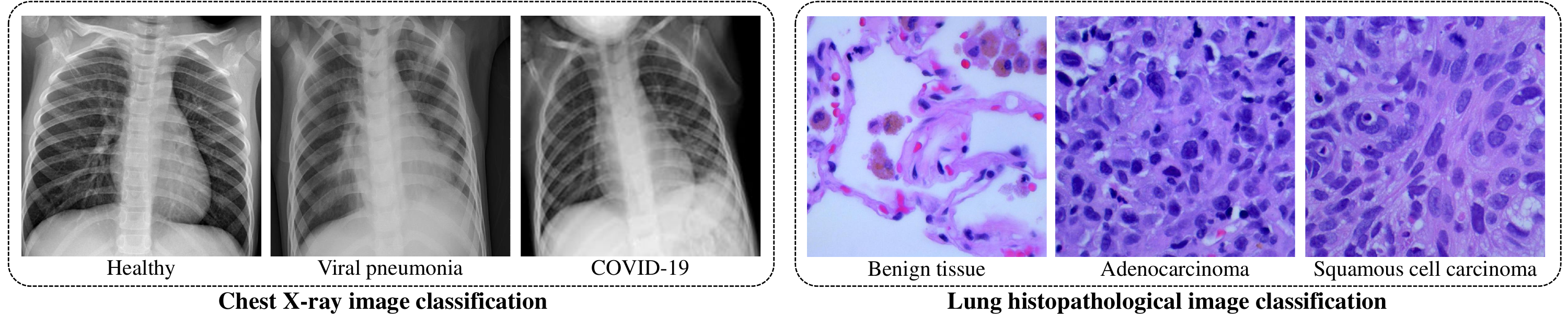}
	\caption{Examples of images used in two few-shot multi-class lung disease diagnosis tasks.}
	\label{fig.two_tasks}
\end{figure*}
\begin{figure}[!t]
	\centering
	\includegraphics[width=0.48\textwidth]{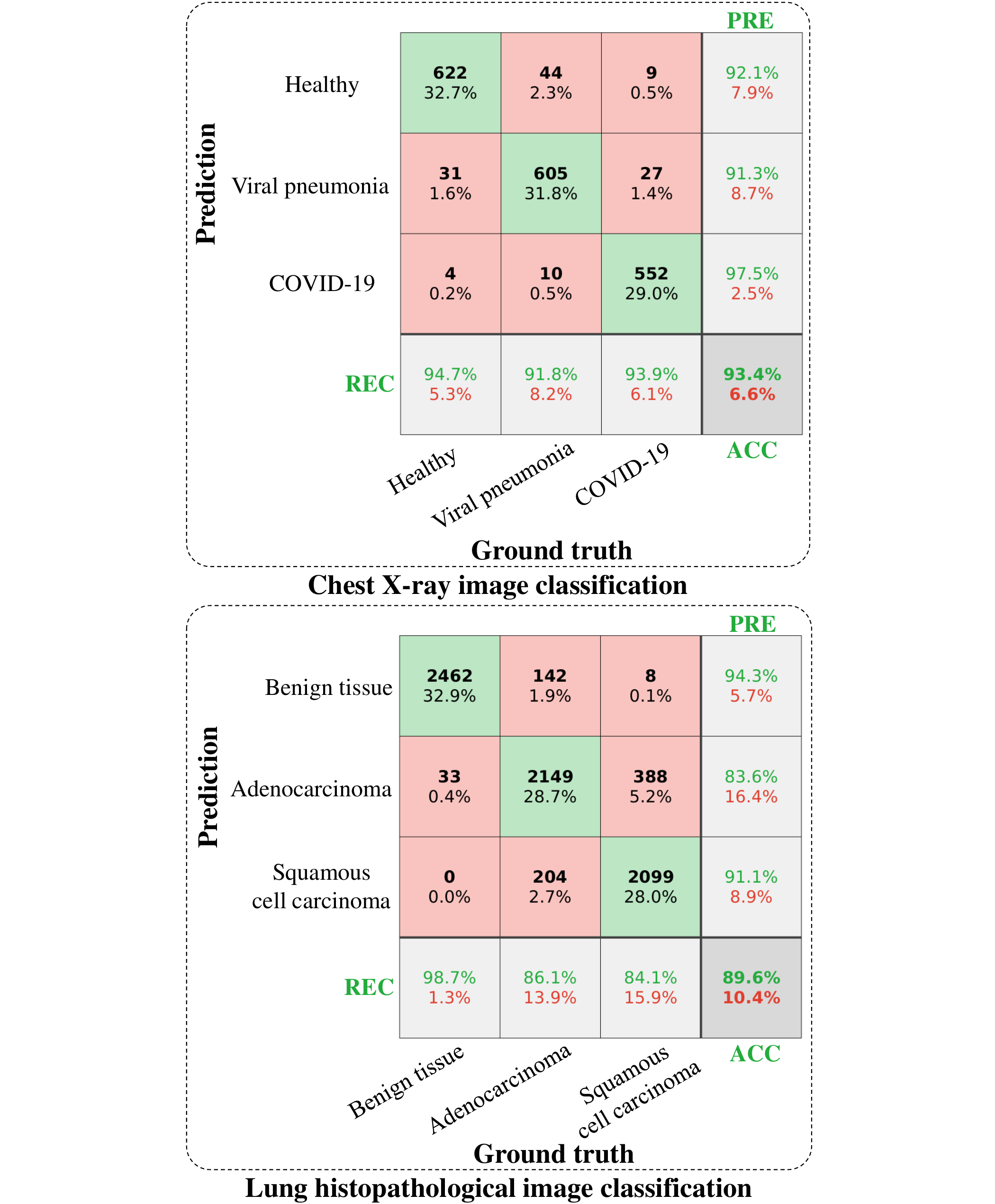}
	\caption{Experimental results of two few-shot lung disease diagnosis tasks. REC: recall; PRE: precision; ACC: accuracy. }
	\label{fig.two_tasks_results}
\end{figure}

We conduct two additional few-shot {multi-class} lung disease diagnosis experiments: (a) chest X-ray image classification for COVID-19 and viral pneumonia detection \cite{chowdhury2020can, rahman2020exploring}, and (b) lung histopathological image classification for lung cancer diagnosis \cite{borkowski2019lung}, to validate the effectiveness of our proposed MTSN and CWCE loss. The first experiment has three classes of images: (1) healthy, (2) viral pneumonia, and (3) COVID-19 (an example of each class is shown in Fig. \ref{fig.two_tasks}), while the second experiment also has three classes of images: (1) benign tissue, (2) adenocarcinoma, and (3) squamous cell carcinoma (an example of each class is shown in Fig. \ref{fig.two_tasks}). In these two experiments, we only select a few images from each class for MTSN training. The numbers of images used for training, validation, and testing are given in Table \ref{tab.two_tasks_image_nums}, where it can be observed that the training set is much smaller than the validation and test sets. Since (\ref{eq.L}) can only be used for binary image classification problems, we extend it here to tackle multi-class image classification problems.

\begin{table}[!t]
	\fontsize{8.2}{10}\selectfont
	\caption{Training, validation, and test sample sizes in the two few-shot biomedical image classification experiments.}  
	\label{tab.two_tasks_image_nums}
	\begin{subtable}[h]{0.45\textwidth}
		\begin{center}
			\begin{tabular}{c|ccc}
				\toprule
				\multicolumn{1}{c}{Class} & Training & Validation & Test  \\
				\hline
				\hline
				Healthy & 27 & 657  & 657 \\
				Viral pneumonia & 27 & 659 & 659 \\
				COVID-19 & 24 & 588 & 588 \\ 
				\bottomrule
			\end{tabular}
			\caption{Chest X-ray image classification.}
		\end{center}
	\end{subtable}
	
	\fontsize{8.2}{10}\selectfont
	\begin{subtable}[h]{0.45\textwidth}
		\begin{center}
			\begin{tabular}{c|ccc}
				\toprule
				\multicolumn{1}{c}{Class} & Training & Validation & Test  \\
				\hline
				\hline
				Benign tissue & 10 & 2,495  & 2,495 \\
				Adenocarcinoma & 10 & 2,495 & 2,495\\ 
				Squamous cell carcinoma & 10 & 2,495 & 2,495 \\
				\bottomrule
			\end{tabular}
			\caption{Lung histopathological image
				classification.}
		\end{center}
	\end{subtable}
\end{table}

Let us consider the chest X-ray image classification task as an example. 
Each image $\mathbf{x}$ is assigned  a pair of two labels $(r,s)$ {with the following values}:
\begin{itemize}
	\item $r=1$ and $s=0$ when $\mathbf{x}$ is a healthy image;
	\item $r=2$ and $s=1$ when $\mathbf{x}$ is a viral pneumonia image;
	\item $r=3$ and $s=3$ when $\mathbf{x}$ is a COVID-19 image.
\end{itemize}
The numbers of healthy images (class 1), viral pneumonia images (class 2), and COVID-19 images (class 3) are denoted as $n_1$, $n_2$, and $n_3$, respectively. The total number of images is $N=n_1+n_2+n_3$.
The weight $\omega{_{\text{cla},e}}$ used in the image classification loss $\mathcal{L}{_{\text{cla}}}$ w.r.t. class $e$ is given by:  
\begin{equation}
	\omega{_{\text{cla},e}} = \frac{N-n_e}{2N}.
\end{equation}
$\sum_{e=1}^{3}\omega{_{\text{cla},e}}=1$. Therefore, $\mathcal{L}{_{\text{cla}}}$  can be written as follows:
\begin{align}
	\begin{split}
		\mathcal{L}{_{\text{cla}}}=
		-\sum_{e=1}^{3} 
		\omega{_{\text{cla},e}}
		\bigg( 
		k_{e,i} \log(p_{e}(\mathbf{x}_i))
		+ k_{e,j} \log(p_{e}(\mathbf{x}_j))
		\bigg),
		\label{eq.L_cla}
	\end{split}
\end{align}
where $p_{e}(\mathbf{x})\in[0,1]$ indicates the probability that $\mathbf{x}$ belongs to class $e$. $\sum_{e=1}^3p_{e}(\mathbf{x})=1$. $k_{e,i}=1$ when $e=r_i$, and $k_{e,i}=0$ otherwise. Given a pair of images $\mathbf{x}_i$ and $\mathbf{x}_j$ with ground-truth labels $(r_i,s_i)$ and $(r_j,s_j)$, respectively, there are four cases:
\begin{itemize}
	\item case 1: $|s_i-s_j|=1$ (healthy v.s. viral pneumonia);
	\item case 2: $|s_i - s_j|=2$ (viral pneumonia v.s. COVID-19);
	\item case 3: $|s_i-s_j|=3$ (COVID-19 v.s. healthy);
	\item case 4: $|s_i-s_j|=0$ (two images are of the same class).
\end{itemize}
The weight $\omega{_{\text{sim},c}}$ used in the image similarity comparison loss $\mathcal{L}{_{\text{sim}}}$ with respect to case $c$ is given by:  
{
	\begin{equation}
		\omega{_{\text{sim},c}}=\begin{cases}
			\frac{1}{3}(1-\frac{2n_1n_2}{N(N-1)}) & \text{if\ \ } c=1
			\\
			\frac{1}{3}(1-\frac{2n_2n_3}{N(N-1)}) & \text{if\ \ } c=2
			\\
			\frac{1}{3}(1-\frac{2n_1n_3}{N(N-1)}) & \text{if\ \ } c=3
			\\
			\frac{1}{3}(1-\frac{\sum_{w=1}^{3}n_w(n_w-1)}{N(N-1)})& \text{if\ \ } c=4
		\end{cases},
	\end{equation}
}
$\sum_{c=1}^{4}\omega{_{\text{sim},c}}=1$. Therefore, $\mathcal{L}{_{\text{sim}}}$  can be written as follows:
\begin{align}
		\mathcal{L}_\text{sim}= 
		-\sum_{c=1}^{4}
		\omega{_{\text{sim},c}}
		h_{c,(i,j)}
		\log (q_c(\mathbf{x}_i,\mathbf{x}_j)),
		\label{eq.L_sim}
\end{align}
where $q_{c}(\mathbf{x}_i,\mathbf{x}_j)\in[0,1]$ indicates the similarity between $\mathbf{x}_i$ and $\mathbf{x}_j$ under case $c$. $\sum_{c=1}^4q_{c}(\mathbf{x}_i,\mathbf{x}_j)=1$. $h_{c,(i,j)}=1$ when $\mathbf{x}_i$ and $\mathbf{x}_j$ belong to case $c$, and $h_{c,(i,j)}=0$ otherwise. The hyper-parameter $\lambda$ is empirically set to $0.3$. 

The experimental results of these two multi-task biomedical image classification tasks are presented in Fig. \ref{fig.two_tasks_results} with two confusion matrices. These results demonstrate that our proposed MTSN {can be effectively trained with very few images to solve multi-class biomedical image classification problems}. {Specifically}, the achieved accuracy values for chest X-ray image classification ($\sim$25-shot learning) and lung histopathological image classification (10-shot learning) are 93\% and 90\%, respectively. The chest X-ray image classification result compares favorably {with the accuracy range of} 82\%-93\% achieved by supervised methods (2,520 images for training, 840 images for validation, and 840 images for testing) using all available training data \cite{jadon2021covid}. Although the accuracy achieved by MTSN for lung histopathological image classification is {lower than the accuracy of over 97\%} reported in \cite{abbas2020histopathological} by supervised approaches using the full training set (8,250 images for training, $\sim$3,000 images for validation, and 3,744 images for testing), we believe that our proposed low-shot learning method can achieve comparable results when a small amount of additional images are incorporated for MTSN training.

\section{Discussion}
\label{sec.discussion}

Extensive experiments demonstrate the effectiveness and efficiency of training an MTSN by minimizing our proposed CWCE loss. Such a low-shot learning approach significantly reduces over-fitting and achieves an accuracy on a small training set (1,147 fundus images) comparable to a large training set (approximately 53K fundus images). We also demonstrate its effectiveness on two additional multi-class few-shot biomedical image classification tasks. Additionally, the MTSNs fine-tuned with OVV self-training outperform the SoTA semi-supervised glaucoma diagnosis algorithms \cite{al2019semi,diaz2019retinal} as well as general-purpose semi-supervised learning algorithms \cite{wang2022freematch,chen2023softmatch,sohn2020fixmatch,zhang2021flexmatch} trained for glaucoma diagnosis. They perform similarly, and in some cases, better than SoTA supervised algorithms. However, our proposed method has two limitations:
\begin{itemize}
\item In the OVV self-training, each target fundus image must be compared with all the reference fundus images in the same mini-batch, {resulting in a computational complexity of $\mathscr{O}(n^2)$}. As the mini-batch size increases, OVV self-training becomes relatively memory-consuming. {The high computational complexity of OVV self-training may reduce the feasibility of this method in clinical practice for now. Therefore, we plan to improve the OVV self-training strategy in the future by adaptively selecting only a limited number of reference fundus photographs for semi-supervised glaucoma diagnosis, which can reduce the computational complexity and make the method more practical in clinical settings.}
\item Our proposed OVV self-training strategy is developed for binary image classification and may not be directly applicable to multi-class image classification problems. Therefore, we plan to extend the contrastive prediction procedure to handle multi-class image classification problems in future work. More hyper-parameter tuning can always be done, but it is so easy to over-fit with limited data.
\end{itemize}

\section{Conclusion}

The main contributions of this paper include: (1) a multi-task Siamese network that can learn glaucoma diagnosis from very limited labeled training data; (2) an effective semi-supervised learning strategy, referred to as One-Vote Veto self-training, which can produce pseudo labels for the unlabeled data to fine-tune a pre-trained multi-task Siamese network. Extensive experiments conducted on four fundus image datasets demonstrated the effectiveness of these proposed techniques. The low-shot learning reduces over-fitting and achieves an accuracy on a small training set comparable to that of a large training set. Furthermore, with One-Vote Veto self-training, the multi-task Siamese networks perform similarly to their backbone CNNs (trained via supervised learning on the full training set) on the OHTS test set and show better generalizability on three additional test sets. The methods introduced in this paper can also be applied to other few-shot multi-class biomedical image classification problems, \eg, COVID-19 and lung cancer diagnosis, and other diseases in which only a small quantity of ground-truth labels are available for network training.

\balance
\normalem
\bibliographystyle{IEEEtran}

\begin{thebibliography}{10}
	\providecommand{\url}[1]{#1}
	\csname url@samestyle\endcsname
	\providecommand{\newblock}{\relax}
	\providecommand{\bibinfo}[2]{#2}
	\providecommand{\BIBentrySTDinterwordspacing}{\spaceskip=0pt\relax}
	\providecommand{\BIBentryALTinterwordstretchfactor}{4}
	\providecommand{\BIBentryALTinterwordspacing}{\spaceskip=\fontdimen2\font plus
		\BIBentryALTinterwordstretchfactor\fontdimen3\font minus
		\fontdimen4\font\relax}
	\providecommand{\BIBforeignlanguage}[2]{{%
			\expandafter\ifx\csname l@#1\endcsname\relax
			\typeout{** WARNING: IEEEtran.bst: No hyphenation pattern has been}%
			\typeout{** loaded for the language `#1'. Using the pattern for}%
			\typeout{** the default language instead.}%
			\else
			\language=\csname l@#1\endcsname
			\fi
			#2}}
	\providecommand{\BIBdecl}{\relax}
	\BIBdecl
	
	\bibitem{weinreb2004primary}
	R.~N. Weinreb and P.~T. Khaw, ``Primary open-angle glaucoma,'' \emph{The
		Lancet}, vol. 363, no. 9422, pp. 1711--1720, 2004.
	
	\bibitem{tham2014global}
	Y.-C. Tham \emph{et~al.}, ``Global prevalence of glaucoma and projections of
	glaucoma burden through 2040: A systematic review and meta-analysis,''
	\emph{Ophthalmology}, vol. 121, no.~11, pp. 2081--2090, 2014.
	
	\bibitem{fan2022detecting}
	R.~Fan \emph{et~al.}, ``Detecting glaucoma in the ocular hypertension study
	using deep learning,'' \emph{JAMA Ophthalmology}, vol. 140, no.~4, pp.
	383--391, 2022.
	
	\bibitem{traverso2005direct}
	C.~Traverso \emph{et~al.}, ``Direct costs of glaucoma and severity of the
	disease: a multinational long term study of resource utilisation in
	{Europe},'' \emph{British Journal of Ophthalmology}, vol.~89, no.~10, pp.
	1245--1249, 2005.
	
	\bibitem{huang2020adverse}
	W.~Huang \emph{et~al.}, ``The adverse impact of glaucoma on psychological
	function and daily physical activity,'' \emph{Journal of Ophthalmology}, vol.
	2020, 2020.
	
	\bibitem{parrish1997visual}
	R.~K. Parrish \emph{et~al.}, ``Visual function and quality of life among
	patients with glaucoma,'' \emph{Archives of Ophthalmology}, vol. 115, no.~11,
	pp. 1447--1455, 1997.
	
	\bibitem{kwon2016association}
	M.~Kwon \emph{et~al.}, ``Association between glaucoma and at--fault motor
	vehicle collision involvement among older drivers: A population-based
	study,'' \emph{Ophthalmology}, vol. 123, no.~1, pp. 109--116, 2016.
	
	\bibitem{mcgwin2015binocular}
	G.~McGwin~Jr \emph{et~al.}, ``Binocular visual field impairment in glaucoma and
	at-fault motor vehicle collisions,'' \emph{Journal of Glaucoma}, vol.~24,
	no.~2, pp. 138--143, 2015.
	
	\bibitem{lag}
	L.~Li \emph{et~al.}, ``Attention based glaucoma detection: A large-scale
	database and cnn model,'' in \emph{Proceedings of the IEEE/CVF Conference on
		Computer Vision and Pattern Recognition (CVPR)}, 2019, pp. 10\,571--10\,580.
	
	\bibitem{fan2023transformer}
	R.~Fan \emph{et~al.}, ``Detecting glaucoma from fundus photographs using deep
	learning without convolutions: Transformer for improved generalization,''
	\emph{Ophthalmology Science}, vol.~3, no.~1, p. 100233, 2023.
	
	\bibitem{ohts1}
	M.~O. Gordon and M.~A. Kass, ``The ocular hypertension treatment study: Design
	and baseline description of the participants,'' \emph{Archives of
		Ophthalmology}, vol. 117, no.~5, pp. 573--583, 1999.
	
	\bibitem{ohts2}
	M.~A. Kass \emph{et~al.}, ``The ocular hypertension treatment study: A
	randomized trial determines that topical ocular hypotensive medication delays
	or prevents the onset of primary open-angle glaucoma,'' \emph{Archives of
		Ophthalmology}, vol. 120, no.~6, pp. 701--713, 2002.
	
	\bibitem{ohts3}
	M.~O. Gordon \emph{et~al.}, ``Assessment of the impact of an endpoint committee
	in the ocular hypertension treatment study,'' \emph{American Journal of
		Ophthalmology}, vol. 199, pp. 193--199, 2019.
	
	\bibitem{li2018efficacy}
	Z.~Li \emph{et~al.}, ``Efficacy of a deep learning system for detecting
	glaucomatous optic neuropathy based on color fundus photographs,''
	\emph{Ophthalmology}, vol. 125, no.~8, pp. 1199--1206, 2018.
	
	\bibitem{gomez2019automatic}
	J.~J. G{\'o}mez-Valverde \emph{et~al.}, ``Automatic glaucoma classification
	using color fundus images based on convolutional neural networks and transfer
	learning,'' \emph{Biomedical Optics Express}, vol.~10, no.~2, pp. 892--913,
	2019.
	
	\bibitem{judy2020automated}
	D.~Judy \emph{et~al.}, ``Automated identification of glaucoma from fundus
	images using deep learning techniques,'' \emph{European Journal of Molecular
		\& Clinical Medicine}, vol.~7, no.~2, pp. 5449--5458, 2020.
	
	\bibitem{serener2019transfer}
	A.~Serener and S.~Serte, ``Transfer learning for early and advanced glaucoma
	detection with convolutional neural networks,'' in \emph{2019 Medical
		Technologies Congress (TIPTEKNO)}.\hskip 1em plus 0.5em minus 0.4em\relax
	IEEE, 2019, pp. 1--4.
	
	\bibitem{thakur2020predicting}
	A.~Thakur \emph{et~al.}, ``Predicting glaucoma before onset using deep
	learning,'' \emph{Ophthalmology Glaucoma}, vol.~3, no.~4, pp. 262--268, 2020.
	
	\bibitem{kim2017few}
	M.~Kim \emph{et~al.}, ``Few-shot learning using a small-sized dataset of
	high-resolution fundus images for glaucoma diagnosis,'' in \emph{Proceedings
		of the 2nd International Workshop on Multimedia for Personal Health and
		Health Care}, 2017, pp. 89--92.
	
	\bibitem{al2019semi}
	M.~Al~Ghamdi \emph{et~al.}, ``Semi-supervised transfer learning for
	convolutional neural networks for glaucoma detection,'' in \emph{2019 IEEE
		International Conference on Acoustics, Speech and Signal Processing
		(ICASSP)}.\hskip 1em plus 0.5em minus 0.4em\relax IEEE, 2019, pp. 3812--3816.
	
	\bibitem{diaz2019retinal}
	A.~D{\'\i}az~Pinto \emph{et~al.}, ``Retinal image synthesis and semi-supervised
	learning for glaucoma assessment,'' \emph{IEEE Transactions on Medical
		Imaging}, vol.~38, no.~9, pp. 2211--2218, 2019.
	
	\bibitem{acrima}
	A.~D{\'\i}az~{Pinto} \emph{et~al.}, ``{CNNs} for automatic glaucoma assessment
	using fundus images: An extensive validation,'' \emph{Biomedical Engineering
		Online}, vol.~18, pp. 1--19, 2019.
	
	\bibitem{digs}
	P.~Sample \emph{et~al.}, ``The {African} descent and glaucoma evaluation study
	({ADAGES}): Design and baseline data,'' \emph{Archives of Ophthalmology},
	vol. 127, no.~9, pp. 1136--1145, 2009.
	
	\bibitem{wang2022freematch}
	Y.~Wang \emph{et~al.}, ``Freematch: Self-adaptive thresholding for
	semi-supervised learning,'' in \emph{the International Conference on Learning
		Representations (ICLR)}, 2023, in press.
	
	\bibitem{chen2023softmatch}
	H.~Chen \emph{et~al.}, ``Softmatch: Addressing the quantity-quality trade-off
	in semi-supervised learning,'' in \emph{the International Conference on
		Learning Representations (ICLR)}, 2023, in press.
	
	\bibitem{sohn2020fixmatch}
	K.~Sohn \emph{et~al.}, ``Fixmatch: Simplifying semi-supervised learning with
	consistency and confidence,'' \emph{Advances in Neural Information Processing
		Systems (NeurIPS)}, vol.~33, pp. 596--608, 2020.
	
	\bibitem{zhang2021flexmatch}
	B.~Zhang \emph{et~al.}, ``Flexmatch: Boosting semi-supervised learning with
	curriculum pseudo labeling,'' \emph{Advances in Neural Information Processing
		Systems (NeurIPS)}, vol.~34, pp. 18\,408--18\,419, 2021.
	
	\bibitem{dosovitskiyimage}
	A.~Dosovitskiy \emph{et~al.}, ``An image is worth 16x16 words: Transformers for
	image recognition at scale,'' in \emph{the International Conference on
		Learning Representations (ICLR)}.
	
	\bibitem{chowdhury2020can}
	M.~E. Chowdhury \emph{et~al.}, ``Can {AI} help in screening viral and
	{COVID-19} pneumonia?'' \emph{IEEE Access}, vol.~8, pp. 132\,665--132\,676,
	2020.
	
	\bibitem{rahman2020exploring}
	T.~Rahman \emph{et~al.}, ``Exploring the effect of image enhancement techniques
	on {COVID-19} detection using chest {X-ray} images,'' \emph{Computers in
		Biology and Medicine}, vol. 132, p. 104319, 2021.
	
	\bibitem{borkowski2019lung}
	A.~A. Borkowski \emph{et~al.}, ``Lung and colon cancer histopathological image
	dataset ({LC25000}),'' \emph{CoRR}, 2019.
	
	\bibitem{alexnet}
	A.~Krizhevsky, ``One weird trick for parallelizing convolutional neural
	networks,'' \emph{CoRR}, 2014.
	
	\bibitem{vgg}
	K.~Simonyan and A.~Zisserman, ``Very deep convolutional networks for
	large-scale image recognition,'' in \emph{the International Conference on
		Learning Representations (ICLR)}, 2015, pp. 1--14.
	
	\bibitem{googlenet}
	C.~Szegedy \emph{et~al.}, ``Going deeper with convolutions,'' in
	\emph{Proceedings of the IEEE Conference on Computer Vision and Pattern
		Recognition (CVPR)}, 2015, pp. 1--9.
	
	\bibitem{inception}
	C.~{Szegedy} \emph{et~al.}, ``Rethinking the inception architecture for
	computer vision,'' in \emph{Proceedings of the IEEE Conference on Computer
		Vision and Pattern Recognition (CVPR)}, 2016, pp. 2818--2826.
	
	\bibitem{ahn2018deep}
	J.~M. Ahn \emph{et~al.}, ``A deep learning model for the detection of both
	advanced and early glaucoma using fundus photography,'' \emph{PloS one},
	vol.~13, no.~11, p. e0207982, 2018.
	
	\bibitem{imagenet_cvpr09}
	J.~Deng \emph{et~al.}, ``{ImageNet: A large-scale hierarchical image
		database},'' in \emph{IEEE Conference on Computer Vision and Pattern
		Recognition (CVPR)}, 2009, pp. 248--255.
	
	\bibitem{resnet}
	K.~He \emph{et~al.}, ``Deep residual learning for image recognition,'' in
	\emph{Proceedings of the IEEE Conference on Computer Vision and Pattern
		Recognition (CVPR)}, 2016, pp. 770--778.
	
	\bibitem{liu2018deep}
	S.~Liu \emph{et~al.}, ``A deep learning-based algorithm identifies glaucomatous
	discs using monoscopic fundus photographs,'' \emph{Ophthalmology Glaucoma},
	vol.~1, no.~1, pp. 15--22, 2018.
	
	\bibitem{ran2019detection}
	A.~R. Ran \emph{et~al.}, ``Detection of glaucomatous optic neuropathy with
	spectral-domain optical coherence tomography: A retrospective training and
	validation deep-learning analysis,'' \emph{The Lancet Digital Health},
	vol.~1, no.~4, pp. e172--e182, 2019.
	
	\bibitem{medeiros2020detection}
	F.~A. Medeiros \emph{et~al.}, ``Detection of progressive glaucomatous optic
	nerve damage on fundus photographs with deep learning,''
	\emph{Ophthalmology}, vol. 128, no.~3, pp. 383--392, 2021.
	
	\bibitem{christopher2018performance}
	M.~Christopher \emph{et~al.}, ``Performance of deep learning architectures and
	transfer learning for detecting glaucomatous optic neuropathy in fundus
	photographs,'' \emph{Scientific Reports}, vol.~8, no.~1, pp. 1--13, 2018.
	
	\bibitem{christopher2020effects}
	M.~{Christopher} \emph{et~al.}, ``Effects of study population, labeling and
	training on glaucoma detection using deep learning algorithms,''
	\emph{Translational Vision Science \& Technology}, vol.~9, no.~2, pp. 27--27,
	2020.
	
	\bibitem{jain2016open}
	D.~Jain \emph{et~al.}, ``Open-source, ultra-low-cost smartphone attachment for
	non-mydriatic fundus photography-open indirect ophthalmoscope,''
	\emph{Investigative Ophthalmology \& Visual Science}, vol.~57, no.~12, pp.
	1685--1685, 2016.
	
	\bibitem{matthew2014smart}
	E.~Matthew~Lawson and R.~Raskar, ``Smart phone administered fundus imaging
	without additional imaging optics,'' \emph{Investigative Ophthalmology \&
		Visual Science}, vol.~55, no.~13, pp. 1609--1609, 2014.
	
	\bibitem{mobilenetv2}
	M.~Sandler \emph{et~al.}, ``{MobileNetv2}: Inverted residuals and linear
	bottlenecks,'' in \emph{Proceedings of the IEEE Conference on Computer Vision
		and Pattern Recognition (CVPR)}, 2018, pp. 4510--4520.
	
	\bibitem{fslsurvey}
	Y.~Wang \emph{et~al.}, ``Generalizing from a few examples: A survey on few-shot
	learning,'' \emph{ACM Computing Surveys (CSUR)}, vol.~53, no.~3, pp. 1--34,
	2020.
	
	\bibitem{zhou2016attention}
	P.~Zhou \emph{et~al.}, ``Attention-based bidirectional long short-term memory
	networks for relation classification,'' in \emph{Proceedings of the 54th
		annual meeting of the Association for Computational Linguistics (volume 2:
		Short Papers)}, 2016, pp. 207--212.
	
	\bibitem{radford2015unsupervised}
	A.~Radford \emph{et~al.}, ``Unsupervised representation learning with deep
	convolutional generative adversarial networks,'' in \emph{the International
		Conference on Learning Representations (ICLR)}, 2015.
	
	\bibitem{selftraining}
	I.~Triguero \emph{et~al.}, ``Self-labeled techniques for semi-supervised
	learning: Taxonomy, software and empirical study,'' \emph{Knowledge and
		Information Systems}, vol.~42, no.~2, pp. 245--284, 2015.
	
	\bibitem{van2020survey}
	J.~E. Van~Engelen and H.~H. Hoos, ``A survey on semi-supervised learning,''
	\emph{Machine Learning}, vol. 109, no.~2, pp. 373--440, 2020.
	
	\bibitem{miller2016key}
	A.~Miller \emph{et~al.}, ``Key-value memory networks for directly reading
	documents,'' in \emph{Proceedings of the 2016 Conference on Empirical Methods
		in Natural Language Processing (EMNLP)}, 2016, pp. 1400--1409.
	
	\bibitem{siamese}
	G.~Koch \emph{et~al.}, ``Siamese neural networks for one-shot image
	recognition,'' in \emph{ICML Deep Learning Workshop}, vol.~2.\hskip 1em plus
	0.5em minus 0.4em\relax Lille, 2015.
	
	\bibitem{jang2021siamese}
	J.~Jang and C.~O. Kim, ``Siamese network-based health representation learning
	and robust reference-based remaining useful life prediction,'' \emph{IEEE
		Transactions on Industrial Informatics}, vol.~18, no.~8, pp. 5264--5274,
	2021.
	
	\bibitem{wang2018learning}
	Q.~Wang \emph{et~al.}, ``Learning attentions: Residual attentional {Siamese}
	network for high performance online visual tracking,'' in \emph{Proceedings
		of the IEEE Conference on Computer Vision and Pattern Recognition (CVPR)},
	2018, pp. 4854--4863.
	
	\bibitem{wang2015face}
	W.~Wang \emph{et~al.}, ``Face recognition based on deep learning,'' in
	\emph{1st International Conference on Human Centered Computing (HCC)}.\hskip
	1em plus 0.5em minus 0.4em\relax Springer, 2015, pp. 812--820.
	
	\bibitem{triplet}
	E.~Hoffer and N.~Ailon, ``Deep metric learning using triplet network,'' in
	\emph{International Workshop on Similarity-Based Pattern Recognition}.\hskip
	1em plus 0.5em minus 0.4em\relax Springer, 2015, pp. 84--92.
	
	\bibitem{khosla2020supervised}
	P.~Khosla \emph{et~al.}, ``Supervised contrastive learning,'' in \emph{Advances
		in Neural Information Processing Systems (NeurIPS)}, vol.~33, 2020, pp.
	18\,661--18\,673.
	
	\bibitem{kendall2018multi}
	A.~Kendall \emph{et~al.}, ``Multi-task learning using uncertainty to weigh
	losses for scene geometry and semantics,'' in \emph{Proceedings of the IEEE
		Conference on Computer Vision and Pattern Recognition (CVPR)}, 2018, pp.
	7482--7491.
	
	\bibitem{mcclosky2006effective}
	D.~McClosky, E.~Charniak, and M.~Johnson, ``Effective self-training for
	parsing,'' in \emph{Proceedings of the Human Language Technology Conference
		of the NAACL, Main Conference}, 2006, pp. 152--159.
	
	\bibitem{sukhbaatar2015end}
	S.~Sukhbaatar, J.~Weston, R.~Fergus \emph{et~al.}, ``End-to-end memory
	networks,'' in \emph{Advances in Neural Information Processing Systems
		(NeurIPS)}, 2015, pp. 2440--2448.
	
	\bibitem{tsne}
	L.~v.~d. Maaten and G.~Hinton, ``Visualizing data using t-{SNE},''
	\emph{Journal of Machine Learning Research}, vol.~9, no.~11, pp. 2579--2605,
	2008.
	
	\bibitem{chen2017rethinking}
	L.-C. Chen \emph{et~al.}, ``Rethinking atrous convolution for semantic image
	segmentation,'' \emph{CoRR}, 2017.
	
	\bibitem{tan2018survey}
	C.~Tan \emph{et~al.}, ``A survey on deep transfer learning,'' in
	\emph{International Conference on Artificial Neural Networks (ICANN)}.\hskip
	1em plus 0.5em minus 0.4em\relax Springer, 2018, pp. 270--279.
	
	\bibitem{densenet}
	G.~Huang \emph{et~al.}, ``Densely connected convolutional networks,'' in
	\emph{Proceedings of the IEEE Conference on Computer Vision and Pattern
		Recognition (CVPR)}, 2017, pp. 4700--4708.
	
	\bibitem{efficientnet}
	M.~Tan and Q.~Le, ``{EfficientNet}: Rethinking model scaling for convolutional
	neural networks,'' in \emph{International Conference on Machine Learning
		(ICML)}.\hskip 1em plus 0.5em minus 0.4em\relax PMLR, 2019, pp. 6105--6114.
	
	\bibitem{gradcamplus}
	A.~Chattopadhay \emph{et~al.}, ``{Grad-CAM}++: Generalized gradient-based
	visual explanations for deep convolutional networks,'' in \emph{2018 IEEE
		Winter Conference on Applications of Computer Vision (WACV)}.\hskip 1em plus
	0.5em minus 0.4em\relax IEEE, 2018, pp. 839--847.
	
	\bibitem{jadon2021covid}
	S.~Jadon, ``{COVID-19} detection from scarce chest x-ray image data using
	few-shot deep learning approach,'' in \emph{Medical Imaging 2021: Imaging
		Informatics for Healthcare, Research, and Applications}, vol. 11601.\hskip
	1em plus 0.5em minus 0.4em\relax International Society for Optics and
	Photonics, 2021, p. 116010X.
	
	\bibitem{abbas2020histopathological}
	M.~A. Abbas \emph{et~al.}, ``The histopathological diagnosis of adenocarcinoma
	\& squamous cells carcinoma of lungs by artificial intelligence: A
	comparative study of convolutional neural networks,'' \emph{MedRxiv}, pp.
	2020--05, 2020.
	
\end{thebibliography}


\end{document}